\documentclass[pdflatex,sn-basic]{sn-jnl}

\usepackage{iftex}
\ifPDFTeX
\usepackage[T2A,T1]{fontenc}
\usepackage[utf8]{inputenc}
\usepackage[russian,english]{babel}
\usepackage{tempora}
\def\cyrdefault#1{\ifx#1\relax\else\DeclareTextSymbolDefault{#1}{T2A}\expandafter\cyrdefault\fi}
\cyrdefault\cyra\CYRA\cyrb\CYRB\cyrv\CYRV\cyrg\CYRG\cyrd\CYRD\cyre\CYRE\cyryo\CYRYO\cyrzh\CYRZH\cyrz\CYRZ\cyri\CYRI\cyrishrt\CYRISHRT\cyrk\CYRK\cyrl\CYRL\cyrm\CYRM\cyrn\CYRN\cyro\CYRO\cyrp\CYRP\cyrr\CYRR\cyrs\CYRS\cyrt\CYRT\cyru\CYRU\cyrf\CYRF\cyrh\CYRH\cyrc\CYRC\cyrch\CYRCH\cyrsh\CYRSH\cyrshch\CYRSHCH\cyrhrdsn\CYRHRDSN\cyrery\CYRERY\cyrsftsn\CYRSFTSN\cyrerev\CYREREV\cyryu\CYRYU\cyrya\CYRYA\relax
\else
\usepackage{fontspec}
\usepackage{polyglossia}
\setmainlanguage{english}
\setotherlanguage{russian}
\newfontfamily\russianfont{FreeSerif.otf}[BoldFont=FreeSerifBold.otf,ItalicFont=FreeSerifItalic.otf,BoldItalicFont=FreeSerifBoldItalic.otf]
\fi

\usepackage{graphicx}
\usepackage{multirow}
\usepackage{url}
\usepackage{adjustbox}
\usepackage{colortbl}
\usepackage{xcolor}
\usepackage{booktabs}
\usepackage{placeins}
\AtBeginDocument{\setlength{\skip\footins}{8pt plus 4pt minus 4pt}}
\AtBeginDocument{\raggedbottom}
\usepackage{expex}
\lingset{everygla=\upshape,everyglb=\itshape,aboveglftskip=.2ex,aboveexskip=1.5ex plus .5ex minus .5ex,belowexskip=1.5ex plus .5ex minus .5ex}
\newcommand{\gab}[1]{{\footnotesize\MakeUppercase{#1}}}
\newsavebox{\fitbox}
\newcommand{\fittable}[2]{%
  \sbox{\fitbox}{\begin{tabular}{#1}#2\end{tabular}}%
  \ifdim\wd\fitbox>\linewidth
    \setlength{\tabcolsep}{2.5pt}%
    \sbox{\fitbox}{\begin{tabular}{#1}#2\end{tabular}}%
  \fi
  \ifdim\wd\fitbox>\linewidth\resizebox{\linewidth}{!}{\usebox{\fitbox}}%
  \else\begin{tabular*}{\linewidth}{@{\extracolsep{\fill}}#1}#2\end{tabular*}\fi}
\newcommand{\landscapetables}{\renewenvironment{table}[1][]{\begin{sidewaystable}[p]}{\end{sidewaystable}}}
\usepackage{lscape}
\usepackage{array}
\usepackage{makecell}
\usepackage{longtable}
\usepackage{amsmath,amssymb,amsfonts}
\usepackage{tikz}
\usetikzlibrary{arrows.meta,calc}

\begin{document}

\title{Hard to Translate, Harder to Detect: Human Analysis Remains Essential in AI Meme Translation}

\author*[1]{\fnm{Olga} \sur{Zamaraeva}}\email{olga.zamaraeva@udc.es}
\author[1]{\fnm{Carla} \sur{Martínez}}\email{carla.gmartinez@udc.es}
\author[1]{\fnm{Carlos} \sur{Gómez-Rodríguez}}\email{carlos.gomez@udc.es}

\affil*[1]{\orgname{Universidade da Coruña, CITIC}, \orgaddress{\city{A Coruña}, \country{Spain}}}

\abstract{We present a dataset and evaluation setup for a problem in creative machine translation: translating culturally loaded passages that carry much of a text's meaning and humor that may depend on subtle or implicit discourse. We study Soviet comedy quotations that function as memes in Russian-language communication. Off-the-shelf difficulty estimators do not find these passages; current models searching the web find most but not all of them, but their own analyses of the difficulty do not help translation as well as a human-validated one. Given a brief human-validated analysis of why each passage is difficult, older models improve modestly; the strongest current translators (GPT-6 Sol and Claude Opus 5.5) improve more, and in human evaluation GPT-6 Sol comes close to the approved references on most films. Newer models have thus become much better at using such an analysis while still not reliably producing it themselves. We release (1) a dataset of five films' Russian subtitles with their audience-selected culturally salient passages, human-validated analyses, and reference translations (into English, Galician, and Spanish), together with every full-film LLM translation and evaluation produced in this study; and (2) an item-level MQM-style evaluation protocol that measures translator and evaluator noise to set the number of runs the evaluation needs, and whose automatic version tracks human judgment of these passages, which reference-free neural metrics do not, whether they score the whole text or the passages alone.}


\maketitle

\section{Introduction}

No one would spontaneously say ``vague doubts torment me'' (\ref{ex:doubts})\footnote{Examples are given in Cyrillic, in scientific transliteration, and with glosses following the Leipzig Glossing Rules, followed by the approved reference translation (\S\ref{sec:hil}) and, where the discussion needs them, translations produced in this study, labeled by translator and condition. Abbreviations: 1, 2, 3 first, second, third person; \gab{acc} accusative; \gab{adj} adjective; \gab{dat} dative; \gab{dim} diminutive; \gab{f} feminine; \gab{fut} future; \gab{gen} genitive; \gab{imp} imperative; \gab{inf} infinitive; \gab{ins} instrumental; \gab{loc} locative; \gab{long} long (attributive) adjective form; \gab{m} masculine; \gab{n} neuter; \gab{neg} negation; \gab{nom} nominative; \gab{pfv} perfective; \gab{pl} plural; \gab{pol} polite; \gab{pst} past; \gab{ptcl} particle; \gab{refl} reflexive; \gab{sg} singular.} about a shaky business deal---the phrase is  too theatrically overwrought for ordinary modern language such as English or Russian. That is exactly why a Russian speaker says it anyway: the line is recognizable on contact as a fifty-year-old film's catchphrase, and dropping its mock-elevated diction onto a mundane worry is what makes the people around them laugh. Soviet comedies from the 1950s--1980s produced a stock of quotations that function today as memes in this way. This makes these films a demanding test case for machine translation: a model can render the surrounding dialogue fluently and still miss the fact that one line is not ordinary dialogue at all, but a culturally loaded unit whose special function in the original film was so important that it has since far surpassed its boundaries. How to translate such things is an open question; effective solutions may range from more literal, intended for the audience well acquainted with the source culture, to cultural adaptations that do not have almost anything in common with the source on the surface but are intended to make the audience laugh or experience some sentiment similar to the one that the Russian audience associates with the meme. In any case, an effective translation should not be confusing or flat.

\ex \label{ex:doubts}%
\begingl
\gla Меня опять терзают смутные сомнения… //
\glb Menja opjat' terzajut smutnye somnenija… //
\glc 1\gab{sg}.\gab{acc} again torment-3\gab{pl} vague-\gab{nom}.\gab{pl} doubt-\gab{nom}.\gab{pl} //
\glft Approved reference: `Vague doubts haunt me once again\ldots' \trailingcitation{(\textit{Ivan Vasilievich})} //
\endgl
\xe

A fluent translation can also flatten a joke that is in plain view. In \textit{Gentlemen of Fortune}, a kindly kindergarten director plays the wolf in the children's show, right after they have sung the Russian version of ``Who's Afraid of the Big Bad Wolf?'', and introduces himself with a rhymed couplet of mock menace (\ref{ex:wolf}), which Russian speakers now quote to present themselves as formidable or highly experienced. Unaided, LLM translators almost always render it accurately but as unrhymed prose, like Opus 5.5's unaided translation in (\ref{ex:wolf}), and the nursery-rhyme comedy is gone. Once told that the line is a meme, the current translators mostly rhyme it, as the same model does given the analysis of \S\ref{sec:hil}. The approved reference does not rhyme: a rhymed rendering that keeps the mock menace is hard to come by, and the human-in-the-loop construction did not arrive at one. That a model offers several is remarkable, and a professional translator would be glad to be handed such a candidate.
\ex \label{ex:wolf}%
\begingl
\gla Я злой и страшный серый волк, я в поросятах знаю толк! //
\glb Ja zloj i strašnyj seryj volk, ja v porosjatax znaju tolk! //
\glc 1\gab{sg} wicked and scary grey wolf 1\gab{sg} in piglet-\gab{pl}.\gab{loc} know-1\gab{sg} sense //
\glft Approved reference: `I'm the Big Bad Wolf, and I know a piglet or two!'\par
Opus 5.5, unaided: `I'm the big bad scary wolf, / I know a thing or two about piglets.'\par
Opus 5.5, with the analysis: `I'm the big bad wolf, so grim and gruff, / and piglets? Oh, I know my stuff!' \trailingcitation{(\textit{Gentlemen of Fortune})} //
\endgl
\xe

As of the time of this writing, commercial LLMs such as GPT, Gemini, and Claude, produce grammatical, natural-sounding translations of routine dialogue as a matter of course. This means, among other things, that evaluating whole-document fluency and adequacy---the concern of most classical MT metrics---is no longer the informative question for creative content. One of the informative questions is: can a system recognize, among thousands of ordinary lines, the small number that carry the cultural and comic weight of the film, and can it explain to itself why they are difficult before attempting to render them? We call the first task \emph{detection} and the second \emph{analysis}, and we treat them as separate empirical questions.

Existing work is not yet fully set up to answer either question, although the interest in the topic is rising. Only a few years ago, it was still typical to assess translation quality by matching the output, sentence by sentence, against reference translations, with the matches aggregated over a test set \citep{papineni2002bleu} or a score predicted for each segment \citep{rei2020comet}, a method whose validity has long been questioned \citep{reiter2018structured} and which is a poor fit for literary content. By now, literary and creative MT research evaluates beyond the sentence, on paragraphs \citep{thai2022par3,wu2024multiref,zhang2025litransproqa} or whole documents \citep{jiang2022blonde}; however, a handful of decisive units in such evaluations are averaged in with thousands of routine units. Furthermore, most metrics, whether adapted to full documents or not, still reward closeness to a reference or general fluency and were not validated for culturally specific, non-literal content: quality-estimation metrics and smaller LLM judges score literal renderings of idioms above idiomatic ones \citep{marmonier2026idiom}, and on literary paragraphs, standard metrics do not agree with expert translators' clear preference for human over machine translation \citep{thai2022par3}. An LLM judge that reads the whole translation seems to inherit the same dilution. The discipline of translation studies has a rich vocabulary for the phenomenon---culture-specific items \citep{aixela1996culture}, creative shifts away from the literal and units of analysis \citep{bayer2011creative}---which has not, to our knowledge, been turned into a reproducible protocol for evaluating LLM translations with quantified uncertainty.

We separate three tasks that prior work tends to conflate: (1) identifying which passages in a text require special cultural treatment, (2) constructing a brief explanation of why each one is difficult, and only then (3) generating and evaluating translations of the full text. To keep the passage selection itself objective, we do not choose the units of analysis ourselves; we take them from an independent, pre-existing audience poll that asked Russian speakers for their favorite quotations from each film. For each such unit, a human-in-the-loop procedure produces an analysis of what is important about the meme and what it is usually taken to convey, and, separately, a human-vetted reference translation, with explicit acceptance criteria. Finally, we score translations with an item-level, reference-grounded MQM protocol \citep{mqm:2014} whose replication design decomposes uncertainty into translation-sampling variance and evaluation variance, so that comparisons between methods come with a quantified confidence interval---closest in spirit to recent work using LLMs to structure MQM annotation at scale \citep{wang2026large}, although our procedure does not require training a new estimator.

We run this protocol on three generations of LLM translators. We find that, with the analysis provided, the older translators gain modestly, the current ones far more, and in human evaluation the strongest translator's meme passages come close to the approved references. Models have become much better at using an analysis; they still need to be given one. Without the analysis, the translators will often flatten what makes the units special. The analysis and reference layer also serves evaluation: an automatic version of our item-level metric tracks human judgment of the culturally loaded passages, while reference-free neural metrics, applied to the whole text or to the passages alone, and a whole-text LLM judge do not.

Soviet-comedy memes are a convenient extreme case. Any translation domain with an identifiable audience that has already told researchers, through fan culture, canon formation, or simply repeated quotation, which passages matter most---other national cinemas, literary catchphrases, scripture, song lyrics, internet meme culture in other languages---admits the same decomposition: select units by reception, construct a human-validated analysis/reference layer for those units specifically, and evaluate with an item-level protocol whose uncertainty is quantified.

\section{Related work}
\label{sec:related-work}

\subsection{Automated translation evaluation}

Automatic MT evaluation has largely developed around similarity to reference translations, from n-gram overlap \citep{papineni2002bleu,popovic2015chrf} and edit distance \citep{snover2006ter} to embedding similarity \citep{zhang2020bertscore}, an assumption long known to be violated in practice \citep{callison2006re,reiter2018structured}; for LLM literary translation, a systematic review finds overlap metrics broadly agreeing with expert or reader judgments in only one of the seven studies that report both \citep{maharianto2026literary}. Learned metrics such as COMET \citep{rei2020comet} and its quality-estimation variant CometKiwi \citep{rei2022comet22} correlate better with human judgments \citep{freitag2022wmt22metrics}, but they still output a single score per segment or document, trained on routine translation quality; none was designed to isolate a few decisive passages within a long, fluent text. Successors such as xCOMET, which also marks error spans \citep{guerreiro2024xcomet}, and MetricX-24 \citep{juraska2024metricx} are among the strongest metrics on LLM output on WMT's general translation test sets \citep{freitag2024wmt24metrics}; whether that holds for culturally loaded creative passages is one of the questions we test (\S\ref{sec:eval-metrics-compared}).

MQM \citep{mqm:2014} replaces a single quality judgment with a declared typology of errors, which later work weights by severity, and large-scale studies such as \citet{freitag2021experts} confirm that expert, context-aware MQM annotation substantially changes system rankings relative to older crowd-sourced Direct Assessment. The cost of trained-expert MQM annotation motivates using LLMs as judges. How far such judges can be trusted is unsettled: agreement with humans varies widely by annotation scheme, language pair, and domain \citep{almaghout2026annotation}, judges may still penalize creative literary solutions \citep{gerrits2026creativity}, quality-estimation metrics prefer literal renderings of idioms \citep{marmonier2026idiom}, and an LLM evaluator tends to favor its own output \citep{xu2025selfbias,chen2025selfpreference}, a bias that studies of LLM literary translation rarely audit \citep{maharianto2026literary}. \citet{wang2026large}, the closest prior work to our evaluation design, simplify MQM annotation to train a COMET-style estimator. We instead keep the LLM as the evaluator and measure its variance.

Concentrating evaluation on hard material is a current direction. As general test sets saturate, the WMT general task now samples documents by estimated difficulty \citep{kocmi2025wmt25findings}, dedicated models estimate how hard a source text is to translate \citep{proietti2025difficulty}, and difficult texts can be generated on purpose \citep{zouhar2025difficult}. These approaches define difficulty by how badly systems translate a text. Our passages are selected by how much they matter to an audience, and we show that difficulty estimators do not find them (\S\ref{sec:detection}). On the statistical side, our replication design follows the argument that model evaluations should report statistical uncertainty, including that from the randomness of the model's outputs \citep{miller2024errorbars}, and that metric differences need to be read against what humans can distinguish \citep{kocmi2024maze}.

\subsection{Human-in-the-loop translation workflows}

Our HIL procedure (\S\ref{sec:hil}) includes an LLM drafting candidate analyses and candidate reference translations for the units of analysis, and a human expert accepting, revising, or rejecting each candidate. This design follows two lines of prior work. First, \citet{zouhar2024evaluating} show that standard reference translations, produced under normal time and effort constraints, systematically understate achievable quality, and that references produced with targeted expert effort reach markedly higher quality. Second, work on LLM-assisted annotation has found that LLM drafts can match or exceed the quality and consistency of unaided human annotation at a fraction of the cost \citep{gilardi2023chatgpt}, which is the efficiency assumption behind having the LLM draft summaries, analyses, and reference candidates for the human to adjudicate.

\subsection{Literary and creative machine translation}

Empirically, literary and long-document translation remain measurably harder for both neural MT and LLMs than the sentence-level, general-domain translation on which most systems are benchmarked. \citet{thai2022par3} and \citet{utsuro2025aozora} build literary parallel corpora, of world literature aligned by paragraph and of Japanese--English literature from Aozora Bunko and Project Gutenberg aligned by sentence, because standard parallel corpora underrepresent literary register and discourse structure. At the level of LLM-based translation itself, \citet{wang2024longtext} show that translation quality degrades as document length grows and that this is not simply fixed by larger context windows. Closer to our setting, \citet{du2025GPTcreativity} compare prompting strategies for creative literary translation with ChatGPT and find that some work better than others, a minimal instruction to translate creatively being the best, but that none brings the output close to human translation in most of the languages studied; \citet{wu2024multiref} show a complementary result, that literary MT systems trained on multiple, meaningfully varied references mostly outperform those trained on a single reference, which suggests that a single reference underrepresents the acceptable translations of creative content (\S\ref{sec:eval}). Finally, dedicated evaluation resources and studies for specific creative phenomena---figurative language \citep{figurativeMT2024}, metaphor \citep{metaphorMT2025}, and literary quality more broadly via professional question-answering \citep{zhang2025litransproqa}---have emerged precisely because general MT metrics were found not to track these phenomena, which is the same gap this paper targets for culturally loaded, humor-bearing passages, and memes. The closest work to ours is a culture-aware MT benchmark built from novels, which finds that models that show the cultural knowledge a passage needs when probed for it often fail to apply it in translation, and that reference translations make LLM judges more reliable \citep{yuan2026canmt}. Our results point the other way for the current models: once told which passages are memes and why, they apply the analysis well, and the harder step is finding the passages and seeing what makes them difficult. The two findings need not conflict, since a probe that asks about a given item does the finding for the model. We add audience-based selection, the separation of detection from analysis, and a comparison across model generations. Related benchmarks bear on which information helps with such units: on Moroccan Darija idioms, metaphors, and proverbs, recognizing the figurative meaning and producing a figurative translation are distinct abilities, and context helps inconsistently \citep{rbaiti2026darija}; for social-media comments, restoring the conversational context helps most models but does not remove culturally ineffective translations \citep{wu2026culturecmt}; and for classical Chinese heritage texts, human-curated explanatory glosses improve LLM translation substantially \citep{yan2026yrc}. For subtitles specifically, LLM translation has been improved with film metadata in the prompt \citep{pramodya2025subtitles} and with multi-agent systems that carry terminology and cultural context across a whole series, evaluated with a subtitle-adapted MQM over sliding windows of whole episodes \citep{jin2026babel}. In film subtitles of culture-specific items, ChatGPT and DeepSeek stay source-oriented where human subtitlers universalize, recreate, or delete \citep{tang2026nezha}.

In translation, the hardest cases are those where a translator must depart from the source to preserve its effect. \citet{bayer2011creative} operationalizes this, defining \emph{translational creativity} by quantifiable ``creative shifts'' away from the literal and \emph{units of analysis} as the segments where such shifts are called for, terminology we adopt for the passages we evaluate.

\subsection{Humor, memes, and culture-specific items in translation}

The quotations we study fall under what \citet{aixela1996culture} terms culture-specific items (CSIs): elements whose meaning or function in the source culture does not necessarily have an established equivalent in the target culture, such that translating them may require a strategic choice (e.g.\ substitution, generalization, or omission). Humor compounds this problem. \citet{attardo1991script}'s General Theory of Verbal Humor models a joke as a bundle of six knowledge resources: script opposition, logical mechanism, situation, target, narrative strategy, and language. In humor translation, the translator must decide case by case whether the comic effect outranks literal content \citep{zabalbeascoa2005humor}. Computationally, the CLEF JOKER track finds that LLMs preserve both the meaning and the wordplay of a pun in only a minority of cases \citep{ermakova2025joker}, and cultural adaptation with LLMs is an active task in its own right \citep{singh2024crosscultural}. When asked to adapt culture-specific entities, LLMs from different families disagree on substitutions and compress cultural diversity \citep{suchdev2026eid}. Meme translation has so far been studied for image memes \citep{zhao2026memes,zheng2026transmeme}; the memes in our data are quotations, recognized from their wording alone. At the same time, their success as memes may well depend on visual or audio co-occurrence that the meme users later simply remember, but there is no obvious way of translating that non-textual content. 

The Soviet-comedy quotations in our dataset are thus memes in the sense developed in internet culture studies: units of popular culture that are circulated, imitated, and transformed by their users, creating a shared cultural experience in the process, where what is imitated may be a text's content, its form, or its stance, that is, the position the speaker takes toward the text and its addressees \citep[p.~367]{shifman2013memes}. This is consistent with the reception-based selection method we use (\S\ref{sec:data}): what makes a passage a meme is not a property of the text alone, but the fact that a speech community continues to reproduce and recognize it.

\section{Dataset}
\label{sec:data}

Soviet comedies have generated a noticeable amount of memes which are still being used in the Russian language. As outlined above, for the purposes of this paper, we define memes as quotations that are commonly used in everyday communication, usually to make the situation less tense and/or more funny. They are immediately recognized by most Russian speakers of several generations. Table \ref{tab:soviet-films-dataset} presents the films we used in this study. All five films contain widely quoted lines, and four of them were released in cinemas outside the former Soviet Union, including in Finland, Hungary, East Germany, Argentina, Japan, and the United States, and some are now on international streaming services such as Netflix and Amazon Prime Video.\footnote{Foreign releases as listed on IMDb and Kinoafisha (accessed November 2026): \textit{Carnival Night} in Japan, Hungary, East Germany, Romania, and the United States (\url{https://www.imdb.com/title/tt0049397/releaseinfo/}); \textit{The Diamond Arm} in Finland, East Germany, Argentina, and Colombia (\url{https://www.imdb.com/title/tt0062759/releaseinfo/}); \textit{Gentlemen of Fortune} in Poland, East Germany, Hungary, and Finland (\url{https://www.imdb.com/title/tt0068519/releaseinfo/}); \textit{Ivan Vasilievich} in Finland, Hungary, Poland, Germany, Sweden, and Argentina (\url{https://www.kinoafisha.info/en/movies/1624187/}). \textit{The Pokrov Gates} was made for Soviet television. Streaming, e.g.\ \textit{The Diamond Arm}: \url{https://www.netflix.com/title/60024887}, \url{https://www.primevideo.com/detail/The-Diamond-Arm/0N02SYTF2LNX0ZGNV1AZSQEEBA}.}

\begin{table}[!htbp]
\centering
\fittable{lllr}{
\toprule
Original title & Year & English title & Subtitle length (GPT tokens) \\
\midrule
Иван Васильевич меняет профессию & 1973 & Ivan Vasilievich Changes Profession & 31,003\\
Бриллиантовая рука & 1969 & The Diamond Arm & 40,430 \\
Джентльмены удачи & 1971 & Gentlemen of Fortune & 19,971 \\
Карнавальная ночь & 1956 & Carnival Night & 22,975 \\
Покровские ворота & 1982 & The Pokrov Gates & 55,658\\
\bottomrule
}
\caption{Films used in the experiments.}
\label{tab:soviet-films-dataset}
\end{table}

An Internet user (username dubikvit in the Livejournal platform) has independently\footnote{The authors are not acquainted with the user and have not communicated with them; the polls were done several years before the present study and are available publicly at \url{https://dubikvit.livejournal.com/746167.html}.} implemented a number of polls related to the most recognizable citations from Soviet comedies. The user, who had hundreds of subscribers, asked them to propose\footnote{Importantly, the poll organizer did not propose any selection; thus the end-result selection reflects the actual popularity of the memes with a degree of objectivity that arises from the large number of audience members proposing the same things independently.} their favorite, most memorable citations from each of the selected films. The organizer then listed the proposed passages with the percentage of total votes corresponding to each. The participants could propose as many citations as they wanted to or could remember. Naturally, the citations that have become memes have gathered a significant proportion of votes.

For each film, we choose a percentage cut-off according to the distribution of votes, between 2 and 5\%, so that the selected passages correspond to a clear discontinuity in quotation popularity. The selection follows the poll results alone. We use these externally identified culturally salient passages as the basis for meme-level evaluation (\S\ref{sec:eval}).

\begin{table}[ht]
\centering
\small
\fittable{lrrrr}{
\toprule
Film & Participants & Total votes & Cut-off (\%) & Memes \\
\midrule
Ivan Vasilievich Changes Profession & 439 & 6,814 & 3 & 11 \\
The Diamond Arm & 521 & 8,150 & 2 & 23 \\
Gentlemen of Fortune & 378 & 3,686 & 2 & 18 \\
Carnival Night & 264 & 1,502 & 5 & 8 \\
The Pokrov Gates & 530 & 4,610 & 2 & 11 \\
\bottomrule
}
\caption{The audience poll behind the meme selection: participants and votes per film, the share of votes a passage needed to be selected, and the number of memes selected.}
\label{tab:eval-data}
\end{table}

\subsection{Released resource}
\label{sec:release}
For each film the release contains the source subtitles, the audience-selected memes with their position in the subtitle file, the human-validated analysis of each meme, the approved reference translations, and the film-level context used in the prompts, as well as the human-in-the-loop construction logs for \textit{Ivan Vasilievich}; and, for every translator and condition, the full translated subtitle file, each meme passage as rendered in it, the translator's own analysis under the corresponding \textit{meme-search} method, and every automatic and human evaluation. The source subtitles come from OpenSubtitles.org, are identified by their file IDs and checksums, and are redistributed, together with their translations, under the OpenSubtitles.org terms of use.

Table \ref{tab:counts} summarizes the released translations and evaluation runs, and Table \ref{tab:tokens} the size of the released subtitle text in tokens, by language pair and in total.

% Generated by code/paper_tables.py; do not edit by hand.
\begin{table}[ht]
\centering
\small
\fittable{llrrrr}{
\toprule
Language pair & Translator & Films & Memes & Translations & Evaluation runs \\
\midrule
Russian$\to$English & GPT-6 Sol & 5 & 71 & 143 & 286 \\
Russian$\to$English & Opus 5.5 & 5 & 71 & 252 & 504 \\
Russian$\to$English & GPT-5.2 & 5 & 71 & 355 & 710 \\
Russian$\to$English & GPT-5.4-mini & 1 & 23 & 31 & 62 \\
Russian$\to$Galician & GPT-6 Sol & 1 & 11 & 18 & 36 \\
Russian$\to$Galician & GPT-5.2 & 1 & 11 & 193 & 386 \\
Russian$\to$Spanish & GPT-5.2 & 1 & 11 & 41 & 82 \\
\bottomrule
}
\caption{Released translations and automatic evaluation runs per language pair and translator. Memes are counted once per film; each translation is a full subtitle file; each evaluation run scores all memes of one translation.}
\label{tab:counts}
\end{table}

% Generated by code/paper_tables.py; do not edit by hand.
\begin{table}[ht]
\centering
\small
\fittable{lrrrr}{
\toprule
Language pair & Films & Source tokens (Russian) & Translations & Translation tokens \\
\midrule
Russian$\to$English & 5 & 170,037 & 781 & 24,502,648 \\
Russian$\to$Galician & 1 & 31,003 & 211 & 5,706,769 \\
Russian$\to$Spanish & 1 & 31,003 & 41 & 1,116,794 \\
\midrule
Total & 5 & 170,037 & 1,033 & 31,326,211 \\
\bottomrule
}
\caption{Size of the released subtitle text in tokens (o200k\_base tokenizer; whole subtitle files, time codes included, as in Table~\ref{tab:soviet-films-dataset}), by language pair. Source tokens count each film once; in the total they are the Russian text, and each pair's translation tokens are the text in its target language.}
\label{tab:tokens}
\end{table}

\subsection{Data split}
\textit{Ivan Vasilievich} (the top  line in Table \ref{tab:eval-data}) was used to develop the translation conditions and the evaluation prompt; these were then fixed and applied unchanged to the other films.

\section{Methods}
\label{sec:methods}

\subsection{Translation conditions}
\label{sec:hypotheses-and-pipelines}
We compare the translation conditions in Table \ref{tab:methods-hypotheses}. Each adds one kind of guidance to a zero-shot prompt; conditions are not combined. \textit{Characters} stands for generic film-level context: an LLM-drafted, human-vetted description of the film's characters. The three meme-list conditions draw on the same curated list of passages (\S\ref{sec:data}) with progressively more human-approved guidance, and \textit{meme-search} replaces the human-curated list with the model's own web search and analysis. In a pilot with GPT-5.2 on one film we also tested four further kinds of film-level context (plot summary, universal-narrative framing, intertextual references, and examples of humor from the film) and a negative control that supplies a different film's character list as if it were correct; none of them behaved differently from \textit{characters} or \textit{zero} (\S\ref{sec:weak}), so they were not carried forward.

Prior work on prompting for MT finds that what matters most in few-shot examples is their quality, while choosing them for relevance to the input helps little and inconsistently \citep{zhang2023prompting,vilar2023prompting}, and that into English, few-shot examples add little over zero-shot translation \citep{hendy2023good}; so we expect generic film-level context to help little, and the models to translate the memes fluently but flatly. Handing the model the list of memes removes the detection problem; adding a brief analysis of why each is difficult also removes the analysis problem; approved reference translations reduce the task to placing a known solution, a ceiling condition (still potentially very useful to translation industry). \textit{Meme-search} tests whether the model can supply detection and analysis itself.

\begin{table}[]
    \centering
    \small
    \begin{tabular}{lp{0.47\textwidth}p{0.29\textwidth}}
        Condition & Added to the zero-shot prompt & Expected effect on Meme-MQM \\
        \hline
        zero & Nothing. & Worst. \\
        characters & LLM-drafted, human-corrected character descriptions. & Little or no improvement. \\
        \hline
        meme list & Curated list of meme passages, with no hints beyond their identity. & Improvement if detection alone is the bottleneck. \\
        list+analysis & Meme list plus a brief human-approved analysis of why each item is difficult. & Further improvement. \\
        given trans. & Meme list, analysis, and human-approved reference translations. & Ceiling. \\
        \hline
        meme-search & The model searches the web for the film's memes and analyzes them itself; no human curation. & Close to \textit{zero} if the model cannot detect and analyze unaided. \\
    \end{tabular}
    \caption{Translation conditions and the expected effect of each on Meme-MQM (\S\ref{sec:meme-mqm}).}
    \label{tab:methods-hypotheses}
\end{table}

\subsection{Human-in-the-loop construction of analysis and reference layers}
\label{sec:hil}
The human-in-the-loop procedure is a data-construction step. For each film, an LLM (Gemini 3 or GPT-5) is used to draft film-level summaries, meme-level analyses, and candidate translations for culturally salient passages. A human expert then accepts, rejects, or revises these artifacts according to explicit criteria: factual correctness with respect to the film, plausibility of the cultural/pragmatic explanation, idiomaticity in the target language, and preservation of comic or stylistic force where possible. The resulting artifacts are then serialized and used as experimental inputs for the translation and evaluation pipelines. For each annotator, the procedure took under one person-day of work per film.

\begin{figure}[tb]
\centering
\resizebox{\textwidth}{!}{%
\begin{tikzpicture}[
  font=\footnotesize,
  >={Stealth[length=1.8mm]},
  box/.style={draw, rounded corners=2pt, align=center, text width=3.6cm, minimum height=1.2cm, inner sep=3pt,
              execute at begin node={\hyphenpenalty=10000\exhyphenpenalty=10000}},
  llm/.style={box, fill=blue!7, draw=blue!50!black},
  hum/.style={box, fill=orange!14, draw=orange!60!black},
  data/.style={draw, align=center, text width=3.6cm, minimum height=0.7cm, fill=gray!10, inner sep=3pt},
  lane/.style={font=\footnotesize\bfseries, align=right, anchor=east},
  feed/.style={->, dashed, gray!70!black},
]
% inputs
\node[data] (subs)  at (2.4,2.0) {Full film subtitles};
\node[data] (memes) at (7.2,2.0) {Meme list (audience poll)};

% LLM drafts
\node[llm] (l1) at (0,0)   {Draft a film summary: plot, characters, themes, intertextual references};
\node[llm] (l2) at (4.8,0) {Explain each meme: why it is funny, salient, or hard to translate};
\node[llm] (l3) at (9.6,0) {Propose translations of each meme into the target language};

% human checks
\node[hum] (h1) at (0,-2.3)   {Correct factual errors, e.g.\ who says what in which scene};
\node[hum] (h2) at (4.8,-2.3) {Accept only a specific, plausible, factually sound analysis};
\node[hum] (h3) at (9.6,-2.3) {Target-language speaker accepts only a funny, punchy, idiomatic line};

% outputs
\node[data] (o1) at (0,-4.2)   {Film summary};
\node[data] (o2) at (4.8,-4.2) {Meme analyses};
\node[data] (o3) at (9.6,-4.2) {Reference translations};

\node[lane] at (-2.05,0)    {LLM\\drafts};
\node[lane] at (-2.05,-2.3) {Human\\decides};

% inputs
\draw[->] (subs.south) -- ++(0,-0.45) -| (l1.north);
\draw[->] (subs.south) -- ++(0,-0.45) -| ([xshift=-6mm]l2.north);
\draw[->] (memes.south) -- ++(0,-0.75) -| ([xshift=6mm]l2.north);
\draw[->] (memes.south) -- ++(0,-0.75) -| (l3.north);

% review
\draw[->, thick] (l1) -- (h1);
\foreach \i in {2,3}{
  \draw[->, thick] ([xshift=-5mm]l\i.south) -- ([xshift=-5mm]h\i.north);
  \draw[<-, thick] ([xshift=5mm]l\i.south) -- ([xshift=5mm]h\i.north);
  \node[font=\scriptsize\itshape, align=center] at ($(l\i.south)!0.5!(h\i.north)$) {iterate};
}

% approved outputs
\foreach \i in {1,2,3} \draw[->] (h\i) -- (o\i);

% each approved layer informs the next step
\draw[feed] (o1.east) -- ++(0.6,0) |- ([yshift=-3mm]l2.west);
\draw[feed] (o2.east) -- ++(0.6,0) |- ([yshift=-3mm]l3.west);
\end{tikzpicture}%
}
\caption{Human-in-the-loop construction of the analysis and reference layers for one film. In each round the LLM drafts and a human decides; each approved layer is passed on to the later rounds (dashed arrows).}
\label{fig:hil}
\end{figure}
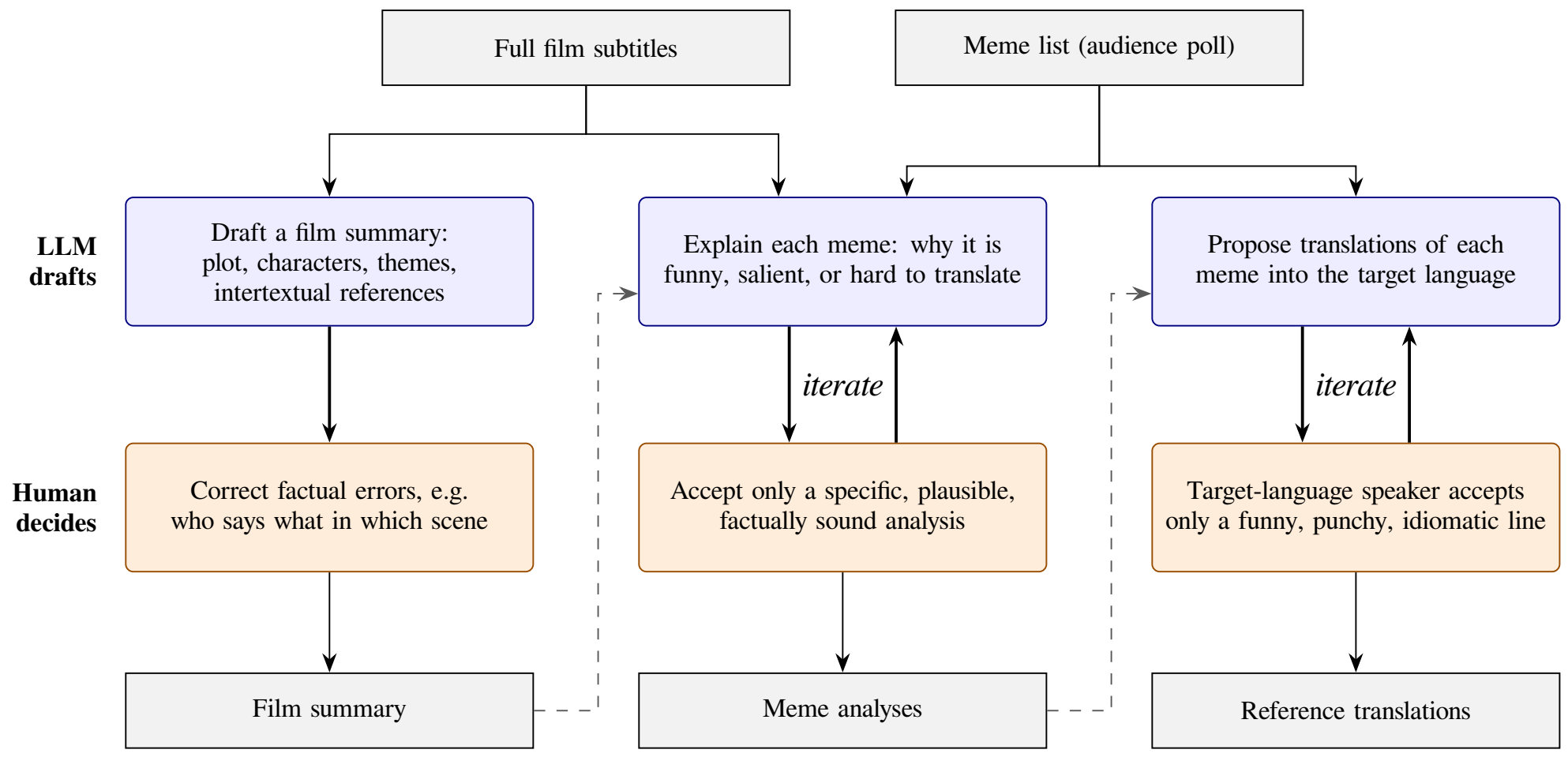

Each film goes through three rounds of the same exchange, in which the LLM drafts and a human decides (Figure~\ref{fig:hil}). First, the LLM reads the full subtitles and writes a structured summary of the film: overview, characters, narrative structure and themes, intertextual references, and moments of high creativity or humor. The human corrects whatever is plainly wrong, such as a line attributed to the wrong character. Second, the LLM explains each meme on the list: why it is funny, culturally salient, pragmatically marked, or otherwise hard to translate. The conversation continues until the analysis says something specific and plausible about the meme and nothing factually wrong. Another expert might approve a different analysis; we assume that any expert-approved analysis is useful downstream. Third, given the summary and the analyses, the LLM proposes translations of each meme, and a fluent speaker of the target language keeps pushing until one is funny, punchy, and idiomatic, a line that would make a foreign viewer laugh without confusing them. Many such translations are possible, and one suffices. These reference translations are not themselves evaluated; they serve the evaluator (\S\ref{sec:eval}). The construction logs for \textit{Ivan Vasilievich} are part of the release (\S\ref{sec:release}).

\paragraph{Annotators.} All HIL steps above, and the spot checks of the segment mapping (\S\ref{sec:mapping}), were carried out by the first two authors. Both have university-level training in philology; the first author also has a certification as a translator in the United States and has worked as a professional translator in the past. The first author, who has native command of Russian and near-native command of English as well as the cultural background needed to judge the plausibility of a proposed analysis, performed the meme-level analysis steps and the Russian-to-English reference-translation steps; the second author, a native speaker of Spanish and Galician, performed the reference-translation and adequacy-judgment steps for Galician and Spanish (\S\ref{sec:pilot}). The human evaluations of translations (\S\ref{sec:eval-metrics-compared}) follow the same division. 

\subsection{Evaluation}
\label{sec:eval}
We evaluate subtitle translations of five Soviet comedy films from Russian into English (and, for one film, into Galician and Spanish), produced by several LLM translators (\S\ref{sec:translation-generation}) under the conditions in Table \ref{tab:methods-hypotheses}. We expect these translations to be broadly fluent and accurate and to differ mainly in how they handle the culturally loaded passages, and we expect metrics that score the translation as a whole to be insensitive to exactly this difference, because the few passages that matter are averaged in with thousands of routine ones.

We compare our item-level metric against two kinds of whole-text evaluation: reference-free neural quality-estimation metrics, and an LLM judge that reads the whole translation.

\subsubsection{Neural quality-estimation metrics}
\label{sec:comet-eval}

We score every translation with three learned metrics in their reference-free (quality-estimation) mode: CometKiwi (wmt22-cometkiwi-da; \citealp{rei2022comet22}), xCOMET-XL \citep{guerreiro2024xcomet}, and MetricX-24 (hybrid-large; \citealp{juraska2024metricx}), the latter two being among the strongest metrics of recent WMT metrics evaluations \citep{freitag2024wmt24metrics}. Segments are subtitle entries, paired by time code between source and translation, and a translation's score is the mean over its entries. For each condition we score the same translations as Meme-MQM and report the mean.

\subsubsection{MQM scoring}
\label{sec:mqm-score}
Both LLM-based measures follow MQM \citep{mqm:2014}: the evaluator lists errors, each with a category (accuracy, fluency, style, or terminology) and a severity. We use only major and minor severity, since in fluent MT output of highly original material we found no meaningful distinction between critical and major errors. With weights $w_{\text{major}}=5$ and $w_{\text{minor}}=1$, the score is
\[
\text{score} = \frac{1}{U}\sum_{i=1}^{N} w_{s(i)},
\]
the number of major-error equivalents per unit, where $U$ is the number of subtitle entries for the whole-text judge and the number of memes for Meme-MQM. A score of 0 means that no errors were found, and a score of 1 that each unit contains, on average, one major error (the score of 1 thus signaling disastrous translation quality).

\subsubsection{LLM whole-text MQM}
\label{sec:llm-wf}
The whole-text judge receives the full subtitle text, source and translation with time codes removed, and lists translation issues following a reference-free MQM-style prompt: major issues harm meaning, humor, character voice, or artistic effect, including cultural references that fail to land, and minor issues are weaknesses in naturalness or style. The prompt tells the judge not to penalize creative solutions that depart from literal meaning but preserve artistic effect. The judge is GPT-5.4 (temperature 1.0) for GPT-5.2 and Claude Sonnet 5.5 for the newer translator models. Scoring per subtitle entry, the natural production unit of subtitles, avoids penalizing long entries differently from short ones.

\subsubsection{Meme-MQM}
\label{sec:meme-mqm}
Meme-MQM scores only the audience-selected memes (\S\ref{sec:data}), a representative though not exhaustive set of the passages on which fluent translations differ. For each translation, the evaluator receives every meme with its source text, the approved reference translation, the approved analysis, and the candidate span, and lists the errors in the candidate, each with a severity (major or minor) and a category (accuracy, fluency, style, or terminology). It is instructed that the candidate need not match the reference, that a different rendering preserving the meaning, tone, humor, and register described in the analysis is not an error, and that a candidate identical to the reference has no errors. It is told to judge what makes the meme work, as the analysis describes it, and not the particular wording, names, or references the analysis mentions: a candidate that replaces a name or reference with another that does the same job for the target audience follows the analysis and is not a mistranslation (Supplementary Material, Section~A). The main evaluator is Claude Sonnet 5.5, called at its default settings through the Claude Code command-line interface with tools disabled, with two evaluation runs per translation. For comparison, every translation was also scored by an evaluator from the other model family, GPT-5-mini for Russian--English and GPT-5.4-mini for the Russian--Galician (\S\ref{sec:pilot}) and Russian--Spanish (\S\ref{sec:pilot-spa}) pilots (reasoning effort ``low'' and ``none'' respectively, no sampling temperature), and the current translators also by Claude Haiku 5.5 (\S\ref{sec:eval-metrics-compared}).

\paragraph{Locating the memes.}\label{sec:mapping} A \emph{segment-mapping} step locates each meme in every candidate translation, so that only the meme passage is scored. Translators are instructed to preserve subtitle time codes, so each source subtitle entry has a translated counterpart with the same time code. Where a meme fills its source entries, its translation is the translated text of exactly those entries. Where the entries also contain other dialogue, an LLM (GPT-6 Sol) trims the translated text of those entries to the part that renders the meme; where the entries do not contain the meme at all (the subtitles paraphrase or reorder the quotation), it searches the three entries on either side. Some translations, mostly from the older translators, do not keep the source time codes (they merge, split, or renumber entries); in these, the LLM locates each meme by content in the whole translated file. A proposed span is used only if it occurs verbatim in the translation. Finally, a second LLM (GPT-5.4) checks every span against the Russian lines it is supposed to translate; spans it rejects are located again by content in the whole translated file, and the authors spot-checked the result. The same procedure is applied to every translator and condition, so that no condition is scored on a tighter or looser span than another.

\subsubsection{Measuring noise and choosing the number of runs}
\label{sec:variance}

A Meme-MQM score varies for two reasons: a translator sampled twice may produce different translations, and an LLM evaluator run twice on the same translation may report different errors. Both can be measured, and together they determine how many translations and evaluation runs a comparison needs: when the difference between two conditions is large relative to the noise, a few suffice, and when it is small, more are needed. Noisier translators and evaluators, and smaller differences, need more runs (\S\ref{sec:noise}).

Let $y_{t,e}$ be the score (major-error equivalents per meme) of translation sample $t$ under evaluation run $e$, for a fixed condition and film. We model
\[
y_{t,e} = \mu + u_t + \varepsilon_{t,e},
\qquad
u_t \sim \mathcal{N}(0, \sigma^2_{\text{trans}}),
\qquad
\varepsilon_{t,e} \sim \mathcal{N}(0, \sigma^2_{\text{eval}}),
\]
where $\sigma^2_{\text{trans}}$ is translation-sampling variance and $\sigma^2_{\text{eval}}$ evaluator variance, both estimated with a random-intercept model. With $T$ translations and $E$ evaluation runs each, the variance of the difference between two conditions' means is
\[
\mathrm{Var}(\bar{y}_A - \bar{y}_B)
\approx
2\left(
\frac{\sigma^2_{\text{trans}}}{T}
+
\frac{\sigma^2_{\text{eval}}}{T E}
\right),
\]
and a difference $\Delta$ is resolved at the 95\% level when $1.96\sqrt{\mathrm{Var}(\bar{y}_A - \bar{y}_B)} \le \Delta$. Given variance estimates from a small calibration run, this fixes the $T$ and $E$ needed to resolve a chosen $\Delta$ for each film, which allows us to compare different translation methods with each other. If method A's score is more than a $\Delta$ ahead or behind method B's score, we can say with certain confidence that A is better or worse than B.  

\subsection{Translation Generation}
\label{sec:translation-generation}

\subsubsection{Models}
The main translator is GPT-6 Sol, run on all five films in English and on \textit{Ivan Vasilievich} in Galician, with three translations per condition; on \textit{Gentlemen of Fortune} and \textit{Carnival Night}, where three did not separate the analysis from its nearest competitors, we added translations until the smallest resolvable difference was about 0.1 (Table \ref{tab:strong-films}). Claude Opus 5.5 serves as a second current translator from a different model family, on all five films in English, also with three translations per condition; for unaided translation, the meme list, and the meme list with analysis, we added translations on every film until the smallest resolvable difference was about 0.1 (Table \ref{tab:opus-films}). The older translators are GPT-5.2, run on all five films in English and on \textit{Ivan Vasilievich} in Galician and Spanish, with between three and eighteen translations per condition, and the smaller GPT-5.4-mini, run on one film (\S\ref{sec:weak}). GPT-5.2, GPT-5.4-mini, and GPT-6 Sol are the three generations of GPT translators we compare (\S\ref{sec:evolution}), the middle one represented only by its smaller model on one film; GPT-6 Sol and Opus 5.5 are the current, state of the art translators.

\subsubsection{Parameters}
GPT-5.2 and GPT-5.4-mini are sampled at temperature 0.8, with an independently randomized seed per translation, so that repeated translations are independent draws. GPT-6 Sol accepts only its default temperature (1.0) and was run with reasoning effort set to ``none''; Opus 5.5 was called at its default settings through the Claude Code command-line interface, with tools disabled except for web search and page fetching under \textit{meme-search}. Differences between translator generations therefore combine the model change with a change in sampling settings, and differences between the two current translators may reflect their reasoning settings as much as the models. For \textit{meme-search}, the translator itself performs the web search and analysis, once per batch of translations: every translation in a batch uses the same analysis. Where translations were added later to narrow an estimate (GPT-5.2 on \textit{Carnival Night}; GPT-6 Sol on \textit{Gentlemen of Fortune} and \textit{Carnival Night}), the added batch ran a search of its own, so for these translators and films the spread of \textit{meme-search} scores also reflects differences between searches. Subtitle files that exceed a model's context are split into chunks at subtitle-entry boundaries, with the same utility for translation and evaluation; most films fit in one chunk.

\subsubsection{Prompts}
Every translation prompt shares a common header---a translation instruction naming the source and target language, followed by an instruction to translate the full subtitle file without stopping and to preserve timecodes exactly---to which method-specific material is appended when the method calls for it (e.g.\ a character-description block for \textit{characters}, the curated meme list for \textit{meme list}, the approved reference translations for \textit{given trans.}). The \textit{zero} condition uses the shared header with nothing appended. For the ``+lang'' conditions in the extended pilot (\S\ref{sec:pilot}), a short language-specific quality addendum (e.g.\ for Galician) is appended after the method-specific material. Full text of the shared header, every method-specific addition, the language addendum, and both evaluation prompts (\S\ref{sec:llm-wf}, \S\ref{sec:meme-mqm}) is given in Supplementary Material, Section~A. For \textit{Carnival Night}, the segment numbers of four memes were corrected after the GPT-5.2 translations had been produced, so the meme lists in those prompts pointed a few subtitle entries away from the memes they quoted; the quoted text itself was correct.

\subsection{AI use}
Claude models (Claude Sonnet 4.6, Claude Sonnet 5, and Claude Opus 5.5, through Claude Code) wrote much of the code for this study, including the translation and evaluation scripts, the segment mapping, the statistical analysis, and the generation of the tables, and assisted in writing the paper, including drafting substantial parts of the text. The authors, particularly the first author, directed this work, elaborated the methods, reviewed the code and its outputs, and edited the paper text thoroughly from start to finish, thus taking responsibility for its every word. The study took 18 months, which, in our opinion, makes conversations with Claude comparable with any other communication an author would be having about their ongoing work with colleagues.

\section{Results}
\label{sec:results}

Every Meme-MQM score is a mean over the translations of a condition and the evaluation runs of each translation. Its 95\% confidence interval combines translation-sampling and evaluator variance (\S\ref{sec:variance}), and it is what we use to compare conditions.

\subsection{Validating the evaluator}
\label{sec:eval-metrics-compared}

\begingroup\landscapetables
% Generated by code/paper_tables.py; do not edit by hand.
\begin{table}[ht]
\centering
\small
\fittable{lrrrrrrrrrrr}{
\toprule
 & & \multicolumn{4}{c}{Whole text} & \multicolumn{3}{c}{Memes only} & \multicolumn{3}{c}{Meme-MQM evaluator} \\
\cmidrule(lr){3-6}\cmidrule(lr){7-9}\cmidrule(lr){10-12}
Film & $N$ & CometKiwi & xCOMET & MetricX & LLM judge & CometKiwi & xCOMET & MetricX & Sonnet & GPT-5-mini & Haiku \\
\midrule
\multicolumn{12}{l}{\textit{GPT-6 Sol}} \\
Ivan Vasilievich & 6 & -0.38 & -0.61 & -0.78 & -0.20 & -0.81 & -0.58 & -0.23 & \textbf{+0.81} & +0.51 & +0.61 \\
Diamond Arm & 6 & +0.03 & -0.20 & -0.14 & -0.09 & -0.54 & +0.43 & -0.71 & \textbf{+0.94} & +0.71 & +0.66 \\
Gentlemen of Fortune & 6 & -0.54 & -0.49 & -0.60 & -0.43 & -0.43 & -0.37 & -0.77 & +0.49 & +0.43 & \textbf{+0.71} \\
Carnival Night & 6 & -0.54 & -0.78 & -0.54 & -0.27 & -0.68 & -0.68 & -0.54 & \textbf{+0.85} & +0.34 & +0.69 \\
Pokrov Gates & 6 & +0.55 & +0.55 & +0.49 & -0.06 & -0.87 & -0.35 & -0.78 & +0.93 & \textbf{+0.97} & +0.93 \\
\midrule
Mean & & -0.18 & -0.30 & -0.31 & -0.21 & -0.67 & -0.31 & -0.61 & \textbf{+0.80} & +0.59 & +0.72 \\
$p$ & & 0.807 & 0.935 & 0.942 & 0.848 & 1.000 & 0.938 & 0.999 & $<$0.001 & $<$0.001 & $<$0.001 \\
\midrule
Mean, no references & & -0.37 & -0.43 & -0.54 & -0.04 & -0.48 & -0.29 & -0.45 & \textbf{+0.67} & +0.28 & +0.53 \\
$p$ & & 0.953 & 0.976 & 0.994 & 0.570 & 0.987 & 0.904 & 0.979 & $<$0.001 & 0.112 & 0.007 \\
\midrule
\multicolumn{12}{l}{\textit{Opus 5.5}} \\
Ivan Vasilievich & 6 & +0.32 & +0.12 & +0.12 & -0.99 & -0.49 & +0.06 & +0.09 & +0.70 & \textbf{+0.75} & +0.70 \\
Diamond Arm & 6 & +0.43 & +0.26 & +0.26 & \textbf{+0.60} & -0.31 & -0.26 & -0.26 & +0.54 & +0.31 & \textbf{+0.60} \\
Gentlemen of Fortune & 6 & -0.60 & +0.03 & -0.09 & -0.43 & -0.83 & -0.77 & -0.94 & +0.83 & +0.77 & \textbf{+1.00} \\
Carnival Night & 6 & -0.15 & -0.15 & -0.33 & -0.39 & -0.33 & -0.15 & +0.09 & \textbf{+0.64} & +0.27 & +0.43 \\
Pokrov Gates & 6 & +0.41 & +0.41 & +0.41 & -0.41 & -0.78 & +0.70 & -0.23 & +0.93 & +0.84 & \textbf{+0.94} \\
\midrule
Mean & & +0.08 & +0.13 & +0.07 & -0.32 & -0.55 & -0.09 & -0.25 & \textbf{+0.73} & +0.59 & \textbf{+0.73} \\
$p$ & & 0.346 & 0.259 & 0.362 & 0.946 & 0.998 & 0.663 & 0.893 & $<$0.001 & $<$0.001 & $<$0.001 \\
\midrule
Mean, no references & & -0.22 & -0.20 & -0.20 & -0.23 & -0.47 & -0.09 & -0.26 & +0.52 & +0.28 & \textbf{+0.53} \\
$p$ & & 0.838 & 0.815 & 0.810 & 0.846 & 0.983 & 0.646 & 0.876 & 0.009 & 0.112 & 0.008 \\
\bottomrule
}
\caption{Agreement of automatic metrics with human evaluation for the current translators, Russian$\to$English: Spearman correlation, across translation conditions. A positive value means agreement with the human ranking; $N$: conditions with a human score and a score from every metric, over which all correlations in the row are computed. CometKiwi, xCOMET and MetricX are reference-free; whole text: mean over the time-code aligned subtitle entries; memes only: mean over the meme passages (each meme's source and the span that renders it, the units the human scored). LLM whole-text judge: Claude Sonnet 5.5. Meme-MQM evaluators: Claude Sonnet 5.5 (main), GPT-5-mini, and Claude Haiku 5.5. Mean: over the films. $p$: one-sided permutation test of the mean, the probability of a mean at least as high when a metric's scores are permuted among the conditions within each film. No references: the same without the given trans.\ condition. Bold: the closest agreement with the human ranking in each row.}
\label{tab:metrics-human}
\end{table}

\endgroup

The comparisons in this paper rest on the automatic Meme-MQM evaluator, so we ask how well it, and the whole-text alternatives, reproduce human judgments of the same translations. The evaluator is not meant to be independent of the human. It is given the analysis and the reference that a human approved, and the question is whether, with that input, an LLM can stand in for the human on the many translations and evaluation runs that a comparison needs, so that the human's time goes into the analysis and into a small sample of judgments that checks the evaluator. Table \ref{tab:metrics-human} correlates the ranking of conditions by the first author with their ranking by each automatic metric. The human scored only the meme passages, so the table shows whether a metric tracks the differences between conditions on these passages, not whether it agrees with a human judgment of the whole translation. The comparison is also one between evaluation with and without human input: the neural metrics run reference-free, while the Meme-MQM evaluators are given the approved reference and analysis. For both state-of-the-art LLM translators (GPT-6 Sol and Claude Opus 5.5), the Meme-MQM evaluators agree with the human ranking on every film, the main evaluator least closely on \textit{Gentlemen of Fortune} for GPT-6 Sol and on \textit{Diamond Arm} for Opus 5.5, and GPT-5-mini least closely on \textit{Carnival Night} for both translators. The whole-text metrics do not agree: the three neural quality-estimation metrics, scored on time-aligned subtitle entries, are near zero or negative on average, and so is the LLM whole-text judge. These metrics are among the strongest in recent WMT evaluations on general translation \citep{freitag2024wmt24metrics}; here, the quality differences between conditions sit in a few dozen culturally loaded passages within thousands of routine subtitle entries, and a whole-text score averages them away. Dilution is not the only reason. Scored on the meme passages alone, the units the human judged, the same three metrics do not agree with the human either (Table \ref{tab:metrics-human}, memes only): for both translators they tend to put unaided translation among the best conditions and the analysis condition among the worst, the reverse of the human ranking. On these passages they prefer the rendering that stays close to the source, which is what the analysis exists to replace, as quality-estimation metrics prefer literal renderings of idioms \citep{marmonier2026idiom}.

With six conditions per film, one film's correlation is weak evidence by itself, so we test the mean over the films with a permutation test that permutes a metric's scores among the conditions within each film (Table \ref{tab:metrics-human}). The agreement of every Meme-MQM evaluator is unlikely to arise by chance, and that of every whole-text metric, and of every neural metric on the meme passages, is compatible with chance. The approved references, which every evaluator puts at the ceiling, account for part of the agreement: without that condition the main evaluator and Haiku still agree with the human beyond chance, and GPT-5-mini does not. The same holds below the level of conditions. On single renderings of single memes, pooled over the films, the scores of all three evaluators correlate with the human's, those of the main evaluator and Haiku more closely than GPT-5-mini's, and leaving out the references changes little (Table \ref{tab:meme-level}); the neural metrics, scored on the same renderings, correlate with the human at most weakly, and less closely than every evaluator. The agreement on single renderings is moderate: most renderings have no human-marked error, and an evaluator's two runs on one rendering are themselves noisy (\S\ref{sec:noise}), which is why we compare conditions by means over memes, translations, and runs.

\begingroup\landscapetables
% Generated by code/paper_tables.py; do not edit by hand.
\begin{table}[ht]
\centering
\small
\fittable{lrrrrrrr}{
\toprule
 & & \multicolumn{3}{c}{Meme-MQM evaluator} & \multicolumn{3}{c}{Neural metric, memes only} \\
\cmidrule(lr){3-5}\cmidrule(lr){6-8}
Translator, conditions & Renderings & Sonnet & GPT-5-mini & Haiku & CometKiwi & xCOMET & MetricX \\
\midrule
GPT-6 Sol, all conditions & 406 & +0.44\,{\scriptsize[+0.32,\,+0.55]} & +0.32\,{\scriptsize[+0.16,\,+0.48]} & +0.43\,{\scriptsize[+0.31,\,+0.54]} & +0.06\,{\scriptsize[-0.08,\,+0.21]} & +0.12\,{\scriptsize[-0.04,\,+0.28]} & +0.08\,{\scriptsize[-0.06,\,+0.22]} \\
GPT-6 Sol, without given trans. & 337 & +0.42\,{\scriptsize[+0.27,\,+0.55]} & +0.30\,{\scriptsize[+0.12,\,+0.47]} & +0.42\,{\scriptsize[+0.28,\,+0.54]} & +0.08\,{\scriptsize[-0.09,\,+0.25]} & +0.14\,{\scriptsize[-0.04,\,+0.31]} & +0.11\,{\scriptsize[-0.04,\,+0.27]} \\
\midrule
Opus 5.5, all conditions & 398 & +0.44\,{\scriptsize[+0.32,\,+0.55]} & +0.26\,{\scriptsize[+0.09,\,+0.43]} & +0.39\,{\scriptsize[+0.24,\,+0.52]} & +0.04\,{\scriptsize[-0.09,\,+0.18]} & +0.20\,{\scriptsize[+0.05,\,+0.33]} & +0.08\,{\scriptsize[-0.07,\,+0.21]} \\
Opus 5.5, without given trans. & 328 & +0.42\,{\scriptsize[+0.28,\,+0.54]} & +0.25\,{\scriptsize[+0.06,\,+0.43]} & +0.37\,{\scriptsize[+0.21,\,+0.52]} & +0.05\,{\scriptsize[-0.10,\,+0.20]} & +0.23\,{\scriptsize[+0.06,\,+0.38]} & +0.08\,{\scriptsize[-0.08,\,+0.23]} \\
\bottomrule
}
\caption{Agreement between the human evaluator and each Meme-MQM evaluator and reference-free neural metric on single renderings, for the current translators, Russian$\to$English: Spearman correlation between the human score and the evaluator's score (mean of its two runs) or the metric's score (oriented so that a positive value means agreement) of the same rendering of the same meme, pooled over the five films, with a 95\% bootstrap interval obtained by resampling memes. Renderings: meme passages with a human score.}
\label{tab:meme-level}
\end{table}

\endgroup

The human and the main automatic evaluator both put the approved references at the ceiling and unaided translation among the worst conditions on most films, and both count errors of register and form that the analysis fixes. On \textit{Diamond Arm}, two memes owe their humor to a nonstandard word form, the long adjective form in (\ref{ex:guilty}) and the ``folk''  infinitive in (\ref{ex:pojtit}); the human evaluator and Sonnet both count every rendering in standard English as a major style error. For GPT-6 Sol, only the analysis, which names the register, yields renderings that keep it; Opus 5.5 keeps it in most conditions. On the same film, both penalize the loss of the rhyme in the lottery slogan (\ref{ex:lottery}) in nearly every condition except the references. Sonnet's blind spot is the allusion: on \textit{Carnival Night}, the drunk lecturer's ``five stars'' (\ref{ex:stars}) alludes to the five-star grade of Soviet cognac, and Sonnet seldom penalizes renderings that lose it, so it does not credit the analysis for restoring it in GPT-6 Sol's translation.

\ex \label{ex:guilty}%
\begingl
\gla Не виноватая я! Он сам пришёл! //
\glb Ne vinovataja ja! On sam prišël! //
\glc \gab{neg} guilty-\gab{f}.\gab{long} 1\gab{sg} 3\gab{sg}.\gab{m} self come.\gab{pst}-\gab{m} //
\glft Approved reference: `Twasn't my fault! He came here on his own!' \trailingcitation{(\textit{Diamond Arm})} //
\endgl
\xe

\ex \label{ex:pojtit}%
\begingl
\gla Нет! На это я пойтить не могу! //
\glb Net! Na èto ja pojtit' ne mogu! //
\glc no on this 1\gab{sg} go.\gab{inf} \gab{neg} can-1\gab{sg} //
\glft Approved reference: `No! I ain't doing no such thing!' \trailingcitation{(\textit{Diamond Arm})} //
\endgl
\xe

\ex \label{ex:lottery}%
\begingl
\gla Кто возьмёт билетов пачку, тот получит… {— Водокачку!} //
\glb Kto voz'mët biletov pačku, tot polučit… {— Vodokačku!} //
\glc who take.\gab{pfv}-3\gab{sg} ticket-\gab{gen}.\gab{pl} pack-\gab{acc} that.one get.\gab{pfv}-3\gab{sg} water.tower-\gab{acc} //
\glft Approved reference: `Buy a whole ticket pack, win\ldots\ a tractor rack!'\par
Opus 5.5, own search: `Buy a whole pack, and you'll win\ldots\ --- A rubbish bin!' \trailingcitation{(\textit{Diamond Arm})} //
\endgl
\xe

\ex \label{ex:stars}%
\begingl
\gla Как мы уже видим, две звёздочки, три звёздочки, четыре звёздочки… Лучше всего, конечно, пять звёздочек! //
\glb Kak my uže vidim, dve zvëzdočki, tri zvëzdočki, četyre zvëzdočki… Lučše vsego, konečno, pjat' zvëzdoček! //
\glc as 1\gab{pl} already see-1\gab{pl} two star.\gab{dim}-\gab{gen} three star.\gab{dim}-\gab{gen} four star.\gab{dim}-\gab{gen} better all-\gab{gen} of.course five star.\gab{dim}-\gab{gen}.\gab{pl} //
\glft Approved reference: `As we can already see, two stars, three stars, four stars\ldots\ But of course five stars is best! [The best cognac in the USSR was known as 5*]' \trailingcitation{(\textit{Carnival Night})} //
\endgl
\xe

The blind spots belong to the evaluator model, not to the protocol. The comparison evaluator from the other model family (GPT-5-mini), given the same analyses and spans, misses all three kinds of error: it counts the standard-English renderings of the two \textit{Diamond Arm} memes at most as minor errors, and never penalizes the lost rhyme or the lost cognac allusion. It therefore finds little or no gain from the analysis for GPT-6 Sol on \textit{Diamond Arm} and \textit{Carnival Night}, and rates the \textit{Diamond Arm} meme passages close to error-free in every condition. Its ranking of conditions still agrees with Sonnet's on most films (Table \ref{tab:judges}), and with the human ranking less closely on average for the current translators (Table \ref{tab:metrics-human}), since both evaluators put the references at the ceiling and unaided translation low; what it misses is the size of the analysis gain on films where such memes are common. A third evaluator, Claude Haiku 5.5, penalizes the register errors nearly as heavily as the human evaluator and partly penalizes the lost allusion, though it seldom penalizes the lost rhyme; its ranking of conditions agrees with the human one less closely than Sonnet's for GPT-6 Sol and as closely for Opus 5.5, and for both more closely than GPT-5-mini's. The evaluators differ greatly in harshness, so absolute scores are comparable only under one evaluator, and the evaluator model is a parameter of the protocol that should be checked against a small human sample, as done here. With one translation per condition and a single human evaluator, the human scores carry no confidence intervals; the human evaluator scored all but one of the repeated items identically.

\begingroup\landscapetables
% Generated by code/paper_tables.py; do not edit by hand.
\begin{table}[ht]
\centering
\small
\fittable{lrrrrrrrrrr}{
\toprule
 & & \multicolumn{4}{c}{Whole text} & \multicolumn{3}{c}{Memes only} & \multicolumn{2}{c}{Meme-MQM evaluator} \\
\cmidrule(lr){3-6}\cmidrule(lr){7-9}\cmidrule(lr){10-11}
Film & $N$ & CometKiwi & xCOMET & MetricX & LLM judge & CometKiwi & xCOMET & MetricX & Sonnet & GPT-5-mini \\
\midrule
Ivan Vasilievich & 11 & +0.42 & +0.12 & -0.27 & -0.12 & -0.14 & -0.89 & -0.16 & \textbf{+0.66} & +0.62 \\
Diamond Arm & 6 & -0.26 & -0.37 & -0.20 & -0.49 & -0.54 & -0.89 & -0.43 & +0.71 & \textbf{+0.83} \\
Gentlemen of Fortune & 6 & -0.83 & -0.77 & -0.94 & +0.77 & -0.26 & -0.37 & -0.60 & \textbf{+0.94} & \textbf{+0.94} \\
Carnival Night & 6 & -0.48 & -0.48 & -0.48 & -0.24 & +0.12 & -0.72 & +0.00 & +0.48 & \textbf{+0.60} \\
Pokrov Gates & 6 & -0.09 & -0.09 & -0.09 & -0.54 & -0.26 & \textbf{+0.03} & -0.77 & \textbf{+0.03} & \textbf{+0.03} \\
Ivan Vasilievich (Galician) & 12 & -0.04 & -0.04 & -0.29 & -0.09 & -0.55 & -0.58 & +0.01 & +0.53 & \textbf{+0.73} \\
\midrule
Mean & & -0.21 & -0.27 & -0.38 & -0.12 & -0.27 & -0.57 & -0.33 & +0.56 & \textbf{+0.62} \\
$p$ & & 0.897 & 0.949 & 0.990 & 0.757 & 0.948 & 1.000 & 0.976 & $<$0.001 & $<$0.001 \\
\midrule
Mean, no references & & +0.12 & +0.03 & -0.04 & +0.10 & +0.15 & -0.41 & -0.03 & +0.31 & \textbf{+0.44} \\
$p$ & & 0.259 & 0.441 & 0.576 & 0.296 & 0.219 & 0.988 & 0.557 & 0.047 & 0.008 \\
\bottomrule
}
\caption{Agreement of automatic metrics with human evaluation for GPT-5.2; $N$, whole text and memes only as in Table \ref{tab:metrics-human}. LLM whole-text judge: GPT-5.4. Meme-MQM evaluators: Claude Sonnet 5.5 (main) and GPT-5-mini (GPT-5.4-mini for Galician). Mean, $p$, and no references as in Table \ref{tab:metrics-human}. Bold: the closest agreement with the human ranking in each row.}
\label{tab:metrics-human-gpt52}
\end{table}

\endgroup

For the older translator, GPT-5.2, the same comparison gives the same picture (Table \ref{tab:metrics-human-gpt52}); the whole-text judge here is GPT-5.4. Over the films, the agreement of the Meme-MQM evaluators is again beyond chance, for the main evaluator only narrowly once the references are left out, and that of the neural metrics, on the whole text or on the meme passages alone, and of the whole-text judge is not. The two clearly positive whole-text agreements on single films are chance. CometKiwi's condition means on \textit{Ivan Vasilievich} differ from one another by less than its scores for different translations of the same condition. The LLM judge's agreement on \textit{Gentlemen of Fortune} rests on six conditions with one translation each, which differ no more than a whole-text judge's scores for different translations of the same condition do elsewhere in our data, and on every other film the same judge disagrees with the human ranking. On \textit{Pokrov Gates} the Meme-MQM evaluators do not agree with the human either, because they score the analysis condition best and the human scores it worst (Table \ref{tab:gpt5.2-by-human-mqm}). The main evaluator agrees with the human slightly less closely on GPT-5.2 than GPT-5-mini does. 

% Generated by code/paper_tables.py; do not edit by hand.
\begin{table}[ht]
\centering
\small
\fittable{lrrrrrr}{
\toprule
Film & zero & characters & meme list & list+analysis & given trans. & added noise \\
\midrule
Diamond Arm & 0.25\,{\scriptsize$\pm$0.18} & \textbf{0.15}\,{\scriptsize$\pm$0.10} & 0.22\,{\scriptsize$\pm$0.09} & 0.17\,{\scriptsize$\pm$0.02} & 0.03\,{\scriptsize$\pm$0.05} & 0.23\,{\scriptsize$\pm$0.06} \\
Gentlemen of Fortune & 0.42\,{\scriptsize$\pm$0.06} & 0.34\,{\scriptsize$\pm$0.00} & 0.33\,{\scriptsize$\pm$0.03} & \textbf{0.20}\,{\scriptsize$\pm$0.10} & 0.06\,{\scriptsize$\pm$0.00} & 0.26\,{\scriptsize$\pm$0.06} \\
Carnival Night & 0.19\,{\scriptsize$\pm$0.02} & 0.20\,{\scriptsize$\pm$0.00} & \textbf{0.16}\,{\scriptsize$\pm$0.02} & 0.19\,{\scriptsize$\pm$0.02} & 0.16\,{\scriptsize$\pm$0.02} & 0.20\,{\scriptsize$\pm$0.00} \\
Pokrov Gates & 0.08\,{\scriptsize$\pm$0.09} & \textbf{0.07}\,{\scriptsize$\pm$0.10} & 0.08\,{\scriptsize$\pm$0.05} & 0.22\,{\scriptsize$\pm$0.03} & 0.00\,{\scriptsize$\pm$0.00} & \textbf{0.07}\,{\scriptsize$\pm$0.13} \\
\bottomrule
}
\caption{Human evaluation of GPT-5.2, Russian$\to$English. MQM adapted score (major-error equivalents per meme; lower is better), mean over translations, with the 95\% method-level confidence interval. Bold: the best condition other than the approved references (given trans.), in each row.}
\label{tab:gpt5.2-by-human-mqm}
\end{table}

In the scores themselves, human evaluation of the GPT-5.2 translations (Tables \ref{tab:pilot} and \ref{tab:gpt5.2-by-human-mqm}) agrees with the automatic evaluator on the ends of the scale: unaided translation is poor, approved references reach the ceiling, and the model's own meme search does not help. It is less consistent in the middle, where the automatic evaluator separates conditions by small margins that the human evaluation, with fewer translations per condition, often does not report. On \textit{Pokrov Gates} the two disagree about the analysis itself: the automatic evaluator scores it as the best condition short of the references (Table \ref{tab:weak-films}), the human as the worst. Most of the human-marked errors fall on two memes. For Margarita Pavlovna's scornful \textit{Kulinar!} (literally `Culinarian!', approved reference `Some cook you are!'), GPT-5.2 sometimes replaces the sarcasm with a direct insult (\textit{You helpless oaf!}, \textit{You kitchen menace!}), which the human evaluator counts as a major accuracy error. Alcoholic Velyurov's pompous disclaimer to Kostik's suggestion to have a drink  (approved reference: `For the record: I am not the one who suggested it!') is rendered with the analysis as  \textit{Just so you know---I didn't suggest it!}, also a major (discourse) error, since Kostik \textit{is} the one who  suggested it, and ``just so you know'' sounds odd. The human evaluator is internally consistent: repeated judgments of the same translation correlate at Pearson $r=0.876$ and Spearman $\rho=0.903$ ($n=11$ repeated items).

Each automatic evaluator shares a model family with some of the translators: Sonnet with Opus 5.5, and the GPT evaluators with the GPT translators. An LLM evaluator can favor its own output \citep{xu2025selfbias,chen2025selfpreference}, and possibly output from its own model family. Our comparisons are between conditions for the same translator, where such a bias applies to every condition alike, and evaluators from the two families rank the conditions similarly for every translator and target language (Table \ref{tab:judges}). Absolute scores of translators from different families are less safe to compare: both automatic evaluators score Opus 5.5's unaided translations better than GPT-6 Sol's on every film, while the human evaluator does so on only one (Table \ref{tab:human-current}). We therefore compare translators by how much the conditions change their scores.

The automatic evaluator judges each candidate against the approved analysis, and the human evaluator is the person who approved it, so either could favor a translation merely for following the analysis, and the analysis condition would then win by construction. The human evaluation was blind to the condition: the evaluator saw each rendering without knowing which prompt had produced it. The judgments also show that following the analysis is not what they reward. On \textit{Pokrov Gates}, the human scored GPT-5.2's translations with the analysis as the worst condition, because with the analysis the model sometimes turned one line's sarcasm into a direct insult and mistranslated another line's disclaimer (Table \ref{tab:gpt5.2-by-human-mqm}). For Opus 5.5 on \textit{Carnival Night}, the human found no gain from the analysis (Table \ref{tab:human-current}). And the human accepts solutions that depart from the analysis: Opus 5.5's substitutions under its own guide, which replace references that the approved analysis keeps, count as effective, and its \textit{meme-search} translations are rated above unaided translation on every film. The automatic evaluator is more exposed to the concern. It scored the \textit{Pokrov Gates} analysis condition as the best, and it accepts only some of the substitutions (see below). Its score is best read as a measure of how well a translation realizes what the approved analysis describes: where the analysis is followed badly, or where a different solution is as good, it can err in the direction of the analysis, and the human sample is what detects this.

% Generated by code/paper_tables.py; do not edit by hand.
\begin{table}[ht]
\centering
\small
\fittable{lrrr}{
\toprule
Film & GPT-5.2 & GPT-6 Sol & Opus 5.5 \\
\midrule
Ivan Vasilievich & +0.77 & +0.89 & +0.77 \\
Diamond Arm & +0.83 & +0.83 & +0.37 \\
Gentlemen of Fortune & +0.83 & +0.89 & +0.94 \\
Carnival Night & +0.71 & +0.37 & +0.54 \\
Pokrov Gates & +0.60 & +0.83 & +0.89 \\
\midrule
Mean & +0.75 & +0.76 & +0.70 \\
\midrule
Ivan Vasilievich (Galician) & +0.70 & +0.83 & -- \\
Ivan Vasilievich (Spanish) & +0.90 & -- & -- \\
\bottomrule
}
\caption{Agreement between the main Meme-MQM evaluator (Claude Sonnet 5.5) and an evaluator from the other model family (GPT-5-mini; GPT-5.4-mini for Galician and Spanish), for each translator: Spearman correlation between their rankings of the translation conditions, over the translations both evaluators scored. Mean: English films. --: translator not run in that language.}
\label{tab:judges}
\end{table}

The whole-text judge's complaints show why it fails. On the GPT-5.2 translations, only a small share of GPT-5.4's complaints concern a meme passage at all, and an LLM meta-evaluation of those against the approved analyses finds most of them unnecessarily harsh and few aligned with the actual difficulty; many penalize renderings identical or close to the approved reference (Supplementary Material, Section~B). Whole-text evaluation thus fails on this material for two reasons that compound: the signal is diluted across the text, and where the judge does look at a meme, it often judges it against the wrong standard. The neural metrics fail for the same two reasons: on the whole text the signal is diluted, and on the meme passages alone, without the reference and analysis, they prefer the literal rendering.

LLM evaluators sometimes normalize a genuinely creative, human-approved solution, flagging it as a defect. Ivan the Terrible's dismissal of Bunsha's wife (\ref{ex:sorrow}) has an approved reference that renders the line near-verbatim---a deliberate choice to preserve the Tsar's archaic, mock-regal register against the modern setting (\S\ref{sec:hil}). The LLM whole-text judge (\S\ref{sec:llm-wf}) flags this exact reference wording, used verbatim by \textit{given trans.}, as a major style error: ``This iconic line is rendered too literally to sound funny in English. The mock-poetic register becomes stilted rather than quotably comic.'' The same line is flagged as a major error on 6 further conditions out of the 11 tested, most citing the same excessive-literalness complaint. This is not a translation error; the meta-evaluation of the judge's complaints (Supplementary Material, Section~B) independently confirms as much for the \textit{given trans.} flag: ``the complaint would also apply to the reference translation\ldots\ so it penalizes a reasonable rendering.'' The judge is not wrong that the line reads as literal; it is wrong that literal is a defect here. The item-level evaluator is less bound to the reference. Following its own guide, Opus 5.5 adapts cultural references for an English audience where the approved analysis keeps the original (\S\ref{sec:detection}): the main evaluator calls its ``Nelson'' for ``Kutuzov'' (\ref{ex:kutuzov}) a legitimate substitution that does the same job (both men were blind in one eye), and objects in writing only to what else the line loses. Its severities do not follow its words: it scores the Nelson renderings more harshly than Kutuzov renderings that lose the same things, and of two translations identical but for the name, it gives only the one with Nelson a major error. At least in this case, then, an evaluator's stated acceptance of a substitution does not carry over to its scores. It is split on the rubbish bin for the lottery's industrial prize (\ref{ex:lottery}). It credits Opus's rhymed version of the teetotaler toast (\ref{ex:teetotal}; ``on someone else's dime, the teetotaller / and the ulcer case both have a fine time''), but it still prefers the analysis's Kolyma to Opus's Siberia (\ref{ex:kolyma}), counting Siberia as a vaguer stand-in for the labor camps, even though Siberia would be much better understood by non-Russian audiences.

\ex \label{ex:sorrow}%
\begingl
\gla Оставь меня, старушка, я в печали. //
\glb Ostav' menja, staruška, ja v pečali. //
\glc leave-\gab{imp} 1\gab{sg}.\gab{acc} old.woman.\gab{dim} 1\gab{sg} in sorrow-\gab{loc} //
\glft Approved reference: `Leave me, old woman, I am in sorrow.' \trailingcitation{(\textit{Ivan Vasilievich})} //
\endgl
\xe

\ex \label{ex:teetotal}%
\begingl
\gla Как говорит наш дорогой шеф, за чужой счёт пьют даже трезвенники… и язвенники! //
\glb Kak govorit naš dorogoj šef, za čužoj sčët p'jut daže trezvenniki… i jazvenniki! //
\glc as say-3\gab{sg} our dear boss for other's account drink-3\gab{pl} even teetotaler-\gab{pl} and ulcer.sufferer-\gab{pl} //
\glft Approved reference: `As our beloved boss says, on someone else's dime, even teetotalers and convalescers drink!'\par
Opus 5.5, own search: `As our dear chief says, on someone else's dime, the teetotaller / and the ulcer case both have a fine time.' \trailingcitation{(\textit{Diamond Arm})} //
\endgl
\xe

\ex \label{ex:kutuzov}%
\begingl
\gla {Дитям —} мороженое, его {бабе —} цветы. Смотры, нэ пэрэпутай!.. Кутузоу! //
\glb {Ditjam —} moroženoe, ego {babe —} cvety. Smotry, nè pèrèputaj!.. Kutuzou! //
\glc child-\gab{dat}.\gab{pl} ice.cream his woman-\gab{dat} flower-\gab{pl} look-\gab{imp} \gab{neg} mix.up-\gab{imp} Kutuzov //
\glft Approved reference: `Ice cream for the young'uns, flowers for his broad. Look now, don't mix 'em up! Kutuzov!'\par
Opus 5.5, own search: `Ice cream for the kiddies, flowers for his missus. Don't mix it up, Nelson!' \trailingcitation{(\textit{Diamond Arm})} //
\endgl
\xe

\ex \label{ex:kolyma}%
\begingl
\gla Ну, будете у нас на Колыме!.. {— Нет,} уж лучше Вы к нам! //
\glb Nu, budete u nas na Kolyme!.. {— Net,} už lučše Vy k nam! //
\glc well be.\gab{fut}-2\gab{pl} at 1\gab{pl} on Kolyma-\gab{loc} no \gab{ptcl} better 2\gab{pl} to 1\gab{pl} //
\glft Approved reference: `Well, if you're ever in Kolyma\ldots\ --- No, better you visit us!'\par
Opus 5.5, own search: `Well, if you're ever up our way in Siberia\ldots\ --- No, you'd better come visit us!' \trailingcitation{(\textit{Diamond Arm})} //
\endgl
\xe

\subsection{Measuring noise and the number of runs}
\label{sec:noise}

% Generated by code/paper_tables.py; do not edit by hand.
\begin{table}[ht]
\centering
\small
\fittable{lrrr}{
\toprule
Translator & Translation SD & Evaluator SD & Method$\times$film cells \\
\midrule
GPT-5.2 & 0.078 & 0.091 & 52 \\
GPT-6 Sol & 0.076 & 0.085 & 30 \\
Opus 5.5 & 0.090 & 0.096 & 30 \\
\bottomrule
}
\caption{Pooled standard deviation of the Meme-MQM score across translations of the same condition (translation sampling) and across repeated evaluation runs of the same translation (evaluator), over all English films and conditions with more than one translation. Evaluator: Claude Sonnet 5.5.}
\label{tab:stability}
\end{table}

Both the translator and the evaluator add noise to a Meme-MQM score, and both can be measured (Table \ref{tab:stability}). Pooled over films and conditions, the three translators are about equally noisy, and for all three the evaluator contributes more variance than translation sampling. How many translations and evaluation runs a comparison needs therefore depends less on the translator than on the size of the difference being compared: the older translator's analysis gains were small relative to the noise and needed many runs to resolve, while the newer translator's large gains on some films are resolved with three translations per condition. The closer comparisons, for GPT-6 Sol on two films and for Opus 5.5 on all five, needed more translations, as many as eighteen per condition, and each results table gives the resulting $\Delta$ per film. The procedure is the same in both cases: measure the two variance components on a small calibration run, then choose the number of translations and evaluation runs that resolves the differences of interest (\S\ref{sec:variance}).

\subsection{Finding the memes}
\label{sec:detection}

% Generated by code/paper_tables.py; do not edit by hand.
\begin{table}[ht]
\centering
\small
\fittable{lrrrrrr}{
\toprule
 & \multicolumn{2}{c}{Sentinel-src} & \multicolumn{2}{c}{CometKiwi} & \multicolumn{2}{c}{LLM rating} \\
\cmidrule(lr){2-3}\cmidrule(lr){4-5}\cmidrule(lr){6-7}
Film (meme / all entries) & AUROC & R@K & AUROC & R@K & AUROC & R@K \\
\midrule
Ivan Vasilievich (17/840) & 0.40 & 0.00 & 0.40 & 0.00 & 0.36 & 0.00 \\
Diamond Arm (55/1429) & 0.52 & 0.04 & 0.60 & 0.04 & 0.56 & 0.05 \\
Gentlemen of Fortune (30/558) & 0.54 & 0.03 & 0.53 & 0.10 & 0.64 & 0.13 \\
Carnival Night (12/562) & 0.41 & 0.08 & 0.53 & 0.00 & 0.38 & 0.00 \\
Pokrov Gates (16/1721) & 0.64 & 0.00 & 0.72 & 0.00 & 0.60 & 0.00 \\
\bottomrule
}
\caption{How well off-the-shelf difficulty signals find the meme entries among all subtitle entries of a film; the row label gives the number of meme entries and of all entries. AUROC: probability that a meme entry is ranked harder than a non-meme entry (0.5 = chance). R@K: share of the memes among the $K$ entries ranked hardest, where $K$ is the film's number of meme entries. Sentinel-src (Sentinel-src-25) and the LLM rating (GPT-5-mini, asked how hard each entry is to translate) judge the source alone. CometKiwi, a quality-estimation metric, needs a translation, so a low score on a zero-shot GPT-5.2 translation counts as a sign of difficulty.}
\label{tab:difficulty}
\end{table}

% Generated by code/paper_tables.py; do not edit by hand.
\begin{table}[ht]
\centering
\small
\fittable{lrrrrrr}{
\toprule
 & \multicolumn{2}{c}{GPT-5.2} & \multicolumn{2}{c}{GPT-6 Sol} & \multicolumn{2}{c}{Opus 5.5} \\
\cmidrule(lr){2-3}\cmidrule(lr){4-5}\cmidrule(lr){6-7}
Film & Found & Analyzed & Found & Analyzed & Found & Analyzed \\
\midrule
Ivan Vasilievich & 4/11 & 2/11 & 9/11 & 6/11 & 11/11 & 8/11 \\
Diamond Arm & 9/23 & 6/23 & 18/23 & 14/23 & 20/23 & 18/23 \\
Gentlemen of Fortune & 6/18 & 4/18 & 12/18 & 9/18 & 12/18 & 11/18 \\
Carnival Night & 3/8 & 1/8 & 5/8 & 3/8 & 6/8 & 5/8 \\
Pokrov Gates & 7/11 & 4/11 & 9/11 & 8/11 & 9/11 & 5/11 \\
\bottomrule
}
\caption{The translator's own web search and analysis (\textit{meme-search}) against the audience-selected memes. Found: memes the model's list discusses, in any language or wording. Analyzed: memes for which the model's notes identify the same essential mechanism or pitfall as the approved analysis. Both judged by Claude Sonnet 5.5. Each cell is the search behind translations 1--3, except GPT-5.2 on \textit{Carnival Night}, which shows the second search (translations 4--8).}
\label{tab:coverage}
\end{table}

As a baseline test, we checked if an off-the-shelf difficulty signal could pick out the passages that need attention. Three such signals rank the memes no better than chance, and on some films worse (Table \ref{tab:difficulty}): a dedicated source-side difficulty estimator \citep{proietti2025difficulty}, a quality-estimation metric applied to a zero-shot translation, and an LLM asked directly how hard each subtitle entry is to translate. Most meme entries are short and lexically plain, so a signal that looks at the words of an entry sees nothing unusual in them; what makes them hard is that the audience recognizes them as quotations, which the text itself does not show. Difficulty sampling, which current benchmarks use to find hard test material \citep{kocmi2025wmt25findings}, would therefore not have selected the passages that matter most to the audience.

A translator that searches the web for the film's memes does better. The current translators find clearly more of them than the older one, and analyze correctly a larger share of those they find (Table \ref{tab:coverage}); still, they miss some memes on almost every film, and their own analysis misses the point of some of those they find. Opus 5.5 finds and correctly analyzes the most memes, and writes longer and more prescriptive guides of its own. According to the automatic evaluator, it helps little: with its own guide, Opus improves on unaided translation only slightly and on two films, and never does better than with the bare meme list (Table \ref{tab:opus-films}). However, human evaluation rates the models' own search higher (Table \ref{tab:human-current}): Opus 5.5's \textit{meme-search} translations beat unaided translation on every film, and on \textit{Diamond Arm} come close to the approved references, and GPT-6 Sol's beat unaided translation on three films. Part of the difference lies in what the guide asks for. Following it, Opus adapts cultural references for an English audience where the approved analysis keeps the original: on \textit{Diamond Arm}, Lyolik's mocking ``Kutuzov'' for the bandaged, one-eyed Gesha (\ref{ex:kutuzov}) becomes ``Nelson'', another famous one-eyed commander; the Kolyma labor camps (\ref{ex:kolyma}) become Siberia; and the lottery's anticlimactic industrial prize (\ref{ex:lottery}) becomes a prize that keeps the rhyme (``Buy a whole pack, and you'll win\ldots\ A rubbish bin!''). The human evaluator counts all three as effective solutions. The main evaluator accepts only some of them (\S\ref{sec:eval-metrics-compared}). This is of course a limitation of the LLM evaluator.

An example that proved particularly difficult for the GPT-5.2 translator is the ``caviar'' line of \emph{Ivan Vasilievich} (\ref{ex:caviar}), a satirical inversion of Soviet-era shortages: real black and red caviar pile up at the tsar's banquet, while a cheap Soviet canned spread, called ``eggplant caviar'' in Russian, is presented as a rare overseas delicacy. The approved analysis warns that an English translation must not call the spread ``caviar'', since English ``eggplant caviar'' reads as a real variety of caviar and will confuse most viewers. Every condition without this warning renders the line as ``\ldots\ Overseas caviar\ldots\ Eggplant!'' or similar in every sampled GPT-5.2 run (48/48), and so do GPT-6 Sol and Opus 5.5 in every translation. With the warning, GPT-5.2 still falls into the trap in 11 of its 14 runs, while the current translators avoid it in every translation (``A delicacy from overseas\ldots\ Eggplant spread!''). Every condition renders the line fluently; the failure is in recognizing that a lexically faithful rendering breaks the joke, and what the newer models have gained is the ability to act on a warning once they are given one (\S\ref{sec:evolution}).

\ex \label{ex:caviar}%
\begingl
\gla Икра чёрная, красная… Да! Заморская икра… баклажанная! //
\glb Ikra čërnaja, krasnaja… Da! Zamorskaja ikra… baklažannaja! //
\glc caviar black-\gab{f} red-\gab{f} yes overseas-\gab{f} caviar eggplant-\gab{adj}-\gab{f} //
\glft Approved reference: `Black caviar, red caviar\ldots\ and yes! Overseas spread\ldots\ eggplant!'\par
GPT-5.2, unaided: `Black caviar, red caviar\ldots\ Yes! Overseas caviar\ldots\ Eggplant!' \trailingcitation{(\textit{Ivan Vasilievich})} //
\endgl
\xe

\subsection{Knowing which passages are memes is not always enough}
\label{sec:identity}

Handing the translator the curated list of memes, with no explanation of why each is difficult, solves the detection problem for it. For GPT-5.2, this helps little (Table \ref{tab:weak-films}): on most films the meme list cannot be told apart from the generic character list, and on a few it is slightly better. The current translators gain more from it, GPT-6 Sol by more than $\Delta$ on \textit{Ivan Vasilievich}, \textit{Gentlemen of Fortune}, and \textit{Pokrov Gates}, and Opus 5.5 on every film (Tables \ref{tab:strong-films} and \ref{tab:opus-films}). Human evaluation puts both current translators close to the references with the meme list on \textit{Pokrov Gates} (the film on which unaided translation is already least penalized), and Opus 5.5 also on \textit{Carnival Night} (Table \ref{tab:human-current}). 

This shows a qualitative change in the current generation of LLM translators. GPT-5.2 renders Troshkin's mock-menacing nursery rhyme (\ref{ex:wolf}), repeated here as (\ref{ex:wolf-rep}), as unrhymed prose in every condition and run. GPT-6 Sol does the same unaided, with rare exceptions, and mostly rhymes it once told that the line is a meme; Opus 5.5 occasionally rhymes it unaided, and always when told it is a meme. With the analysis, it produces the rhymed translation in (\ref{ex:wolf-rep}). Given the meme list, GPT-6 Sol renders the stock announcement that the show is over (\ref{ex:kina}), with the sign-off of the Looney Tunes cartoons in every run (``That's all, folks!''), while GPT-5.2 is literal in every condition. Neither solution is in the approved references, and both are the kind of option a translator would be glad to be offered (\S\ref{sec:discussion}). They remain exceptions: on most films, the meme list moves GPT-6 Sol's scores much less than the analysis does (\S\ref{sec:analysis}).

\ex \label{ex:wolf-rep}%
\begingl
\gla Я злой и страшный серый волк, я в поросятах знаю толк! //
\glb Ja zloj i strašnyj seryj volk, ja v porosjatax znaju tolk! //
\glc 1\gab{sg} wicked and scary grey wolf 1\gab{sg} in piglet-\gab{pl}.\gab{loc} know-1\gab{sg} sense //
\glft Approved reference: `I'm the Big Bad Wolf, and I know a piglet or two!'\par
Opus 5.5, unaided: `I'm the big bad scary wolf, / I know a thing or two about piglets.'\par
Opus 5.5, with the analysis: `I'm the big bad wolf, so grim and gruff, / and piglets? Oh, I know my stuff!' \trailingcitation{(\textit{Gentlemen of Fortune})} //
\endgl
\xe
\ex \label{ex:kina}%
\begingl
\gla Всё! Кина не будет. Электричество кончилось. //
\glb Vsë! Kina ne budet. Èlektričestvo končilos'. //
\glc all film.\gab{gen} \gab{neg} be.\gab{fut}.3\gab{sg} electricity end.\gab{pst}.\gab{n}-\gab{refl} //
\glft Approved reference: `That's it! No movie. The electricity's all gone.'\par
GPT-6 Sol, meme list: `That's all, folks! The power's run out.' \trailingcitation{(\textit{Gentlemen of Fortune})} //
\endgl
\xe

\subsection{An analysis helps most}
\label{sec:analysis}
\label{sec:strong}

% Generated by code/paper_tables.py; do not edit by hand.
\begin{table}[ht]
\centering
\small
\fittable{lrrrrrrrr}{
\toprule
Film & $\Delta$ & $T$ & zero & characters & meme list & list+analysis & meme-search & given trans. \\
\midrule
Ivan Vasilievich & 0.16 & 3 & 0.62\,{\scriptsize$\pm$0.15} & 0.54\,{\scriptsize$\pm$0.12} & 0.39\,{\scriptsize$\pm$0.15} & \textbf{0.09}\,{\scriptsize$\pm$0.04} & 0.55\,{\scriptsize$\pm$0.06} & 0.06\,{\scriptsize$\pm$0.06} \\
Diamond Arm & 0.07 & 3 & 0.48\,{\scriptsize$\pm$0.05} & 0.51\,{\scriptsize$\pm$0.04} & 0.44\,{\scriptsize$\pm$0.05} & \textbf{0.31}\,{\scriptsize$\pm$0.03} & 0.43\,{\scriptsize$\pm$0.06} & 0.03\,{\scriptsize$\pm$0.03} \\
Gentlemen of Fortune & 0.09 & 8 & 0.80\,{\scriptsize$\pm$0.09} & 0.80\,{\scriptsize$\pm$0.08} & 0.61\,{\scriptsize$\pm$0.04} & \textbf{0.50}\,{\scriptsize$\pm$0.05} & 0.64\,{\scriptsize$\pm$0.04} & 0.02\,{\scriptsize$\pm$0.02} \\
Carnival Night & 0.12 & 5 & 0.47\,{\scriptsize$\pm$0.04} & 0.42\,{\scriptsize$\pm$0.08} & 0.56\,{\scriptsize$\pm$0.05} & \textbf{0.37}\,{\scriptsize$\pm$0.07} & 0.55\,{\scriptsize$\pm$0.04} & 0.02\,{\scriptsize$\pm$0.04} \\
Pokrov Gates & 0.14 & 3 & 0.50\,{\scriptsize$\pm$0.06} & 0.44\,{\scriptsize$\pm$0.08} & \textbf{0.24}\,{\scriptsize$\pm$0.15} & 0.30\,{\scriptsize$\pm$0.09} & 0.46\,{\scriptsize$\pm$0.02} & 0.01\,{\scriptsize$\pm$0.01} \\
\bottomrule
}
\caption{GPT-6 Sol, Russian$\to$English, evaluator Claude Sonnet 5.5. MQM adapted score (major-error equivalents per meme; lower is better), mean over translations and evaluation runs, with the 95\% method-level confidence interval. $\Delta$: the smallest difference between two conditions with $T$ translations each that is resolved at the 95\% level, with the variances pooled over the film's conditions; $T$: translations of list+analysis. On \textit{Gentlemen of Fortune} every condition has $T$ translations; on \textit{{Carnival Night}} so does every condition but zero, which has 23. Bold: the best condition other than the approved references (given trans.), in each row.}
\label{tab:strong-films}
\end{table}

% Generated by code/paper_tables.py; do not edit by hand.
\begin{table}[ht]
\centering
\small
\fittable{lrrrrrrrr}{
\toprule
Film & $\Delta$ & $T$ & zero & characters & meme list & list+analysis & meme-search & given trans. \\
\midrule
Ivan Vasilievich & 0.06 & 10 & 0.35\,{\scriptsize$\pm$0.04} & 0.35\,{\scriptsize$\pm$0.04} & 0.22\,{\scriptsize$\pm$0.02} & \textbf{0.06}\,{\scriptsize$\pm$0.06} & 0.35\,{\scriptsize$\pm$0.09} & 0.00\,{\scriptsize$\pm$0.00} \\
Diamond Arm & 0.08 & 7 & 0.40\,{\scriptsize$\pm$0.04} & 0.30\,{\scriptsize$\pm$0.12} & 0.25\,{\scriptsize$\pm$0.05} & \textbf{0.20}\,{\scriptsize$\pm$0.06} & 0.32\,{\scriptsize$\pm$0.04} & 0.02\,{\scriptsize$\pm$0.04} \\
Gentlemen of Fortune & 0.06 & 18 & 0.62\,{\scriptsize$\pm$0.04} & 0.65\,{\scriptsize$\pm$0.09} & 0.51\,{\scriptsize$\pm$0.05} & \textbf{0.40}\,{\scriptsize$\pm$0.04} & 0.55\,{\scriptsize$\pm$0.04} & 0.00\,{\scriptsize$\pm$0.01} \\
Carnival Night & 0.07 & 16 & 0.45\,{\scriptsize$\pm$0.06} & 0.38\,{\scriptsize$\pm$0.22} & 0.29\,{\scriptsize$\pm$0.04} & \textbf{0.28}\,{\scriptsize$\pm$0.04} & 0.49\,{\scriptsize$\pm$0.12} & 0.01\,{\scriptsize$\pm$0.02} \\
Pokrov Gates & 0.08 & 18 & 0.45\,{\scriptsize$\pm$0.05} & 0.40\,{\scriptsize$\pm$0.15} & \textbf{0.33}\,{\scriptsize$\pm$0.05} & \textbf{0.33}\,{\scriptsize$\pm$0.05} & 0.47\,{\scriptsize$\pm$0.20} & 0.02\,{\scriptsize$\pm$0.00} \\
\bottomrule
}
\caption{Claude Opus 5.5, Russian$\to$English, evaluator Claude Sonnet 5.5. MQM adapted score (major-error equivalents per meme; lower is better), mean over translations and evaluation runs, with the 95\% method-level confidence interval. $\Delta$: the smallest difference between two conditions with $T$ translations each that is resolved at the 95\% level, with the variances pooled over the film's conditions; $T$: translations of list+analysis. Zero and the meme list have $T$ translations as well; characters and meme-search have three. Bold: the best condition other than the approved references (given trans.), in each row.}
\label{tab:opus-films}
\end{table}

% Generated by code/paper_tables.py; do not edit by hand.
\begin{table}[ht]
\centering
\small
\fittable{lrrrrrr}{
\toprule
Film & zero & characters & meme list & list+analysis & meme-search & given trans. \\
\midrule
\multicolumn{7}{l}{\textit{GPT-6 Sol}} \\
Ivan Vasilievich & 0.40 & 0.33 & 0.30 & \textbf{0.02} & 0.30 & 0.00 \\
Diamond Arm & 0.17 & 0.30 & 0.24 & \textbf{0.15} & 0.28 & 0.01 \\
Gentlemen of Fortune & 0.43 & 0.24 & 0.23 & \textbf{0.02} & 0.29 & 0.09 \\
Carnival Night & 0.15 & 0.15 & 0.15 & \textbf{0.05} & 0.15 & 0.03 \\
Pokrov Gates & 0.12 & 0.04 & \textbf{0.00} & 0.04 & 0.05 & 0.00 \\
\midrule
\multicolumn{7}{l}{\textit{Opus 5.5}} \\
Ivan Vasilievich & 0.62 & 0.24 & 0.25 & \textbf{0.00} & 0.28 & 0.00 \\
Diamond Arm & 0.31 & 0.08 & 0.13 & 0.09 & \textbf{0.05} & 0.01 \\
Gentlemen of Fortune & 0.27 & 0.29 & 0.22 & \textbf{0.13} & 0.21 & 0.03 \\
Carnival Night & 0.15 & 0.15 & \textbf{0.05} & 0.15 & 0.12 & 0.03 \\
Pokrov Gates & 0.14 & \textbf{0.02} & \textbf{0.02} & \textbf{0.02} & 0.12 & 0.00 \\
\bottomrule
}
\caption{Human Meme-MQM (major-error equivalents per meme; lower is better) for the current translators, Russian$\to$English, one translation per condition, one evaluator, blind to the translator and the condition. Agreement with the automatic metrics is in Table \ref{tab:metrics-human}. Bold: the best condition other than the approved references (given trans.), in each row.}
\label{tab:human-current}
\end{table}

With a human-validated analysis, the strongest translator improves on every film, and on three of the five it does better than with any other condition short of the references by more than the film's $\Delta$ (Table \ref{tab:strong-films}); on \textit{Pokrov Gates} the bare meme list does about as well, and on \textit{Carnival Night} the generic character list does, both within $\Delta$. By the automatic evaluator, it comes close to the approved references only on \textit{Ivan Vasilievich} and in Galician (\S\ref{sec:languages}); on the other films it closes roughly a quarter to two fifths of the gap, least on \textit{Carnival Night}. Human evaluation finds larger gains (Table \ref{tab:human-current}): with the analysis, GPT-6 Sol comes close to the approved references on four of the five films, all but \textit{Diamond Arm}. Claude Opus 5.5, a current translator from a different model family, also improves with the analysis on every film by more than $\Delta$ (Table \ref{tab:opus-films}). It already gains more from the bare meme list (\S\ref{sec:identity}), and the analysis adds to the list by more than $\Delta$ only on \textit{Ivan Vasilievich} and \textit{Gentlemen of Fortune}; on the other three films the two conditions cannot be told apart. By the human evaluator, Opus with the analysis comes close to the references only on \textit{Ivan Vasilievich} and \textit{Pokrov Gates} (where generic context does as well), and on \textit{Carnival Night} it does no better than unaided translation. There the two evaluators disagree about the same translations: the automatic evaluator scores the unaided translation the human judged among the worst Opus produced for that film, and the translation with the analysis well above it. The approved references reach the ceiling for every translator on every film.

Looking at individual memes explains most of this. The large gains are concentrated on memes with a concrete discourse ``trap'' that the model falls into unaided and that the analysis names: on \textit{Ivan Vasilievich}, the elliptical line (\ref{ex:doubts-b}), which the model renders as possession although the joke is that the items were stolen; Bunsha's tangled self-introduction as tsar (\ref{ex:tsar}); the list of stolen goods; on \textit{Gentlemen of Fortune}, Fedya's chess advice, which calls the knight a horse and swears to the move with a prison oath (\ref{ex:horse}). For these, unaided translation is penalized heavily in every run and translation with the analysis almost never. On the films with smaller gains, the memes with such a trap are fewer, and the gain is spread thinly over the rest. The approved references remain the most reliable condition on every film. Their few errors, on \textit{Ivan Vasilievich}, come from placement: handed a reference for a line that repeats itself almost word for word (\ref{ex:cure}), GPT-6 Sol sometimes collapses the repetition and shifts the remaining lines into the wrong subtitle entries, perhaps treating the repetition as an error.

\ex \label{ex:doubts-b}%
\begingl
\gla {У Шпака —} магнитофон, {у посла —} медальон. //
\glb {U Špaka —} magnitofon, {u posla —} medal'on. //
\glc {at Shpak-\gab{gen}} tape.recorder {at ambassador-\gab{gen}} medallion //
\glft Approved reference: `Shpak's tape recorder, the ambassador's medallion.'\par
GPT-5.2, unaided: `Shpak has a tape recorder, / the envoy has a medallion.' \trailingcitation{(\textit{Ivan Vasilievich})} //
\endgl
\xe

\ex \label{ex:tsar}%
\begingl
\gla Здравствуйте, царь, очень приятно. Царь, очень приятно, царь. Очень приятно, здравствуйте, царь. //
\glb Zdravstvujte, car', očen' prijatno. Car', očen' prijatno, car'. Očen' prijatno, zdravstvujte, car'. //
\glc hello-\gab{pol} tsar very pleasant tsar very pleasant tsar very pleasant hello-\gab{pol} tsar //
\glft Approved reference: `Hello, I'm the Tsar, very pleased to meet you. The Tsar, very pleased. Tsar, very pleased. Hello, the Tsar.' \trailingcitation{(\textit{Ivan Vasilievich})} //
\endgl
\xe

\ex \label{ex:horse}%
\begingl
\gla Лошадью ходи, век воли не видать! //
\glb Lošad'ju xodi, vek voli ne vidat'! //
\glc horse-\gab{ins} move-\gab{imp} century freedom-\gab{gen} \gab{neg} see-\gab{inf} //
\glft Approved reference: `Move the horse, I'll never see the outside if I'm wrong!' \trailingcitation{(\textit{Gentlemen of Fortune})} //
\endgl
\xe

\ex \label{ex:cure}%
\begingl
\gla И тебя вылечат… и тебя тоже вылечат… и меня вылечат… //
\glb I tebja vylečat… i tebja tože vylečat… i menja vylečat… //
\glc and 2\gab{sg}.\gab{acc} cure.\gab{pfv}-3\gab{pl} and 2\gab{sg}.\gab{acc} too cure.\gab{pfv}-3\gab{pl} and 1\gab{sg}.\gab{acc} cure.\gab{pfv}-3\gab{pl} //
\glft Approved reference: `And they'll cure you\ldots\ and they'll cure you too\ldots\ and they'll cure me\ldots' \trailingcitation{(\textit{Ivan Vasilievich})} //
\endgl
\xe

The clearest case of an analysis put to use is the monument dialogue in \textit{Gentlemen of Fortune}, which ends on a pun. Asked whether the statue is seated (\ref{ex:monument-a}), one of the convicts hears the verb `sit' in its sense `do time' and retorts that nobody would put a monument away (\ref{ex:monument-b}). The analysis states that English cannot reproduce the root play and that one way of compensating it is the ambiguity around being `inside'. Given the analysis, GPT-6 Sol builds the pun around `inside' in all three translations (``Was he inside?'' \ldots\ ``Who's gonna put him inside? He's a monument!''), in its own wording for the rest of the dialogue. Unaided or with only the meme list, it asks whether the man is seated, and the joke is lost. GPT-5.2 and Opus 5.5, given the same analysis, also ask whether he is seated.

\pex \label{ex:monument}%
\a \label{ex:monument-a}%
\begingl
\gla {— Сидячий?} {— Чего?} {— Сидит?} {— Кто?} {— Мужик} этот твой! //
\glb {— Sidjačij?} {— Čego?} {— Sidit?} {— Kto?} {— Mužik} ètot tvoj! //
\glc seated-\gab{m} what sit-3\gab{sg} who man this your //
\glft Approved reference: `Was he inside? --- Who? --- That guy of yours!' \trailingcitation{(\textit{Gentlemen of Fortune})} //
\endgl
\a \label{ex:monument-b}%
\begingl
\gla Кто же его посадит? Он же памятник! //
\glb Kto že ego posadit? On že pamjatnik! //
\glc who \gab{ptcl} 3\gab{sg}.\gab{m}.\gab{acc} seat.\gab{pfv}-3\gab{sg} 3\gab{sg}.\gab{m} \gab{ptcl} monument //
\glft Approved reference: `Who'd put him inside? He's a monument!'\par
GPT-6 Sol, with the analysis: `Who's gonna put him inside? / He's a monument!' \trailingcitation{(\textit{Gentlemen of Fortune})} //
\endgl
\xe

\FloatBarrier
\subsection{Across model generations}
\label{sec:evolution}
\label{sec:model-scale}
\label{sec:weak}
\label{sec:pilot-eng}
\label{sec:cross-film}

% Generated by code/paper_tables.py; do not edit by hand.
\begin{table}[ht]
\centering
\small
\fittable{lrrrrrrrr}{
\toprule
Film & $\Delta$ & $T$ & zero & characters & meme list & list+analysis & meme-search & given trans. \\
\midrule
Ivan Vasilievich & 0.07 & 14 & 0.57\,{\scriptsize$\pm$0.02} & 0.59\,{\scriptsize$\pm$0.07} & \textbf{0.51}\,{\scriptsize$\pm$0.06} & 0.52\,{\scriptsize$\pm$0.08} & 0.59\,{\scriptsize$\pm$0.04} & 0.03\,{\scriptsize$\pm$0.02} \\
Diamond Arm & 0.06 & 4 & 0.47\,{\scriptsize$\pm$0.03} & 0.45\,{\scriptsize$\pm$0.04} & \textbf{0.41}\,{\scriptsize$\pm$0.03} & 0.44\,{\scriptsize$\pm$0.04} & 0.43\,{\scriptsize$\pm$0.02} & 0.01\,{\scriptsize$\pm$0.00} \\
Gentlemen of Fortune & 0.14 & 3 & 0.81\,{\scriptsize$\pm$0.07} & 0.77\,{\scriptsize$\pm$0.14} & 0.68\,{\scriptsize$\pm$0.05} & \textbf{0.63}\,{\scriptsize$\pm$0.10} & 0.67\,{\scriptsize$\pm$0.05} & 0.11\,{\scriptsize$\pm$0.02} \\
Carnival Night & 0.06 & 15 & 0.51\,{\scriptsize$\pm$0.04} & 0.47\,{\scriptsize$\pm$0.04} & 0.49\,{\scriptsize$\pm$0.04} & \textbf{0.38}\,{\scriptsize$\pm$0.05} & 0.57\,{\scriptsize$\pm$0.04} & 0.17\,{\scriptsize$\pm$0.05} \\
Pokrov Gates & 0.07 & 15 & 0.57\,{\scriptsize$\pm$0.06} & 0.56\,{\scriptsize$\pm$0.06} & 0.57\,{\scriptsize$\pm$0.03} & \textbf{0.43}\,{\scriptsize$\pm$0.05} & 0.62\,{\scriptsize$\pm$0.10} & 0.03\,{\scriptsize$\pm$0.04} \\
\bottomrule
}
\caption{GPT-5.2, Russian$\to$English, evaluator Claude Sonnet 5.5. MQM adapted score (major-error equivalents per meme; lower is better), mean over translations and evaluation runs, with the 95\% method-level confidence interval. $\Delta$: the smallest difference between two conditions with $T$ translations each that is resolved at the 95\% level, with the variances pooled over the film's conditions; $T$: translations of list+analysis. The other conditions have between three and 21 translations. Bold: the best condition other than the approved references (given trans.), in each row.}
\label{tab:weak-films}
\end{table}

% Generated by code/paper_tables.py; do not edit by hand.
\begin{table}[ht]
\centering
\small
\fittable{lrrrrrrrrrrr}{
\toprule
Evaluation & \rotatebox{60}{zero} & \rotatebox{60}{summary} & \rotatebox{60}{characters} & \rotatebox{60}{narratives} & \rotatebox{60}{intertext} & \rotatebox{60}{examples} & \rotatebox{60}{meme list} & \rotatebox{60}{list+analysis} & \rotatebox{60}{meme-search} & \rotatebox{60}{given trans.} & \rotatebox{60}{added noise} \\
\midrule
Automatic & 0.57 & 0.52 & 0.59 & 0.61 & 0.54 & 0.58 & \textbf{0.51} & 0.52 & 0.59 & 0.03 & 0.59 \\
 & {\scriptsize$\pm$0.02} & {\scriptsize$\pm$0.02} & {\scriptsize$\pm$0.07} & {\scriptsize$\pm$0.08} & {\scriptsize$\pm$0.07} & {\scriptsize$\pm$0.03} & {\scriptsize$\pm$0.06} & {\scriptsize$\pm$0.08} & {\scriptsize$\pm$0.04} & {\scriptsize$\pm$0.02} & {\scriptsize$\pm$0.06} \\[2pt]
Human & 0.61 & 0.69 & 0.76 & 0.76 & 0.82 & 0.70 & \textbf{0.45} & 0.52 & 0.63 & 0.00 & 0.55 \\
 & {\scriptsize$\pm$0.08} & {\scriptsize$\pm$0.06} & {\scriptsize$\pm$0.14} & {\scriptsize$\pm$0.13} & {\scriptsize$\pm$0.08} & {\scriptsize$\pm$0.07} & {\scriptsize$\pm$0.08} & {\scriptsize$\pm$0.25} & {\scriptsize$\pm$0.03} & {\scriptsize$\pm$0.00} & {\scriptsize$\pm$0.10} \\[2pt]
\bottomrule
}
\caption{Pilot: all eleven conditions on \textit{Ivan Vasilievich}, GPT-5.2, Russian$\to$English. Automatic: evaluator Claude Sonnet 5.5, mean over translations and evaluation runs; human: mean over the translations scored by hand. MQM adapted score (major-error equivalents per meme; lower is better), with the 95\% method-level confidence interval below it. Bold: the best condition other than the approved references (given trans.), in each row.}
\label{tab:pilot}
\end{table}

\begingroup\landscapetables
% Generated by code/paper_tables.py; do not edit by hand.
\begin{table}[ht]
\centering
\small
\fittable{lrrrrrr}{
\toprule
 & \multicolumn{3}{c}{Automatic} & \multicolumn{3}{c}{Human} \\
\cmidrule(lr){2-4}\cmidrule(lr){5-7}
Translator & meme list & list+analysis & meme-search & meme list & list+analysis & meme-search \\
\midrule
\multicolumn{7}{l}{\textit{All five films}} \\
GPT-5.2 & +0.06\,{\scriptsize[+0.02,\,+0.11]} & +0.10\,{\scriptsize[+0.05,\,+0.16]} & +0.03\,{\scriptsize[-0.02,\,+0.08]} & +0.07\,{\scriptsize[-0.01,\,+0.15]} & +0.06\,{\scriptsize[-0.05,\,+0.17]} & -- \\
GPT-6 Sol & +0.13\,{\scriptsize[+0.06,\,+0.20]} & +0.25\,{\scriptsize[+0.17,\,+0.33]} & +0.06\,{\scriptsize[+0.01,\,+0.12]} & +0.06\,{\scriptsize[-0.02,\,+0.15]} & +0.19\,{\scriptsize[+0.08,\,+0.31]} & +0.03\,{\scriptsize[-0.05,\,+0.11]} \\
Opus 5.5 & +0.13\,{\scriptsize[+0.08,\,+0.19]} & +0.20\,{\scriptsize[+0.15,\,+0.27]} & +0.04\,{\scriptsize[-0.02,\,+0.10]} & +0.15\,{\scriptsize[+0.05,\,+0.26]} & +0.21\,{\scriptsize[+0.11,\,+0.32]} & +0.17\,{\scriptsize[+0.06,\,+0.28]} \\
\midrule
\multicolumn{7}{l}{\textit{Without the development film}} \\
GPT-5.2 & +0.07\,{\scriptsize[+0.02,\,+0.11]} & +0.11\,{\scriptsize[+0.05,\,+0.17]} & +0.04\,{\scriptsize[-0.02,\,+0.10]} & +0.06\,{\scriptsize[-0.01,\,+0.15]} & +0.07\,{\scriptsize[-0.05,\,+0.20]} & +0.03\,{\scriptsize[-0.07,\,+0.14]} \\
GPT-6 Sol & +0.10\,{\scriptsize[+0.04,\,+0.18]} & +0.20\,{\scriptsize[+0.12,\,+0.28]} & +0.06\,{\scriptsize[+0.01,\,+0.12]} & +0.05\,{\scriptsize[-0.04,\,+0.15]} & +0.16\,{\scriptsize[+0.04,\,+0.28]} & +0.01\,{\scriptsize[-0.07,\,+0.11]} \\
Opus 5.5 & +0.13\,{\scriptsize[+0.08,\,+0.19]} & +0.19\,{\scriptsize[+0.13,\,+0.25]} & +0.04\,{\scriptsize[-0.03,\,+0.11]} & +0.12\,{\scriptsize[+0.02,\,+0.24]} & +0.15\,{\scriptsize[+0.06,\,+0.25]} & +0.14\,{\scriptsize[+0.03,\,+0.26]} \\
\bottomrule
}
\caption{Gain of each condition over unaided translation, Russian$\to$English, pooled over films: mean over the memes of the difference between the meme's score in unaided translation and in the condition (major-error equivalents per meme; positive: the condition is better), each score averaged over the translations of the condition, with a 95\% bootstrap interval obtained by resampling memes. Automatic: Claude Sonnet 5.5, all translations; human: the translations scored by hand. \textit{Ivan Vasilievich} is the film on which the conditions and the evaluation prompt were developed. --: the condition was not scored on every film.}
\label{tab:gain}
\end{table}

\endgroup

Across three generations of translators, two things change little and one changes a great deal. Unaided translation of the memes barely improves (Tables \ref{tab:weak-films} and \ref{tab:strong-films}). The newer model's own search finds more of the memes (\S\ref{sec:detection}), and translation with its own list and analysis improves little. What changes is the use of a human-validated analysis. The smaller GPT-5.4-mini, run on \textit{Diamond Arm}, gains nothing from it, and GPT-5.2 gains modestly and noisily (Tables \ref{tab:weak-films} and \ref{tab:pilot}). GPT-6 Sol gains more from it than GPT-5.2 on two of the five films and in Galician, by far the most on \textit{Ivan Vasilievich}, and about as much on the other three (Tables \ref{tab:weak-films}, \ref{tab:strong-films}, and \ref{tab:galician}). Human evaluation shows the difference on more films: of the five films on which both generations were evaluated by a human, GPT-6 Sol with the analysis comes close to the approved references on four, and GPT-5.2 with the analysis on none (Tables \ref{tab:pilot}, \ref{tab:gpt5.2-by-human-mqm}, and \ref{tab:human-current}). Pooling the films and pairing the conditions by meme gives each translator's gain an interval (Table \ref{tab:gain}). Under the automatic evaluator, the gain from the analysis is positive for all three translators and larger for both current ones than for GPT-5.2; under the human evaluator, it is distinguishable from zero for the current translators only. \textit{Ivan Vasilievich}, the film on which the conditions were developed, contributes GPT-6 Sol's largest gain: without it the current translators' estimated gains are still larger than GPT-5.2's, but their intervals under the automatic evaluator overlap with GPT-5.2's. The table also shows the two evaluators' disagreement about the models' own search: the human finds a gain from it for Opus 5.5 that the automatic evaluator does not (\S\ref{sec:detection}). A single meme shows the change most plainly (\S\ref{sec:detection}): warned by the analysis not to call an eggplant spread `caviar', GPT-5.2 still does so in most runs, while both current translators follow the warning in every translation. The newer models have become much better at using an expert's analysis; their ability to find and analyze the memes themselves has improved less, and is not yet reliable.

For GPT-5.2, which we tested first and most extensively, the pilot on \textit{Ivan Vasilievich} covered eleven conditions (Table \ref{tab:pilot}). The five kinds of film-level context differ little from one another, and a negative control that presents a different film's character list as correct is not reliably worse than unaided translation, consistent with evidence from classification and multiple-choice tasks that the content of in-context examples matters less than their format \citep{min2022rethinking}.

\subsection{Other target languages}
\label{sec:languages}
\label{sec:pilot}
\label{sec:pilot-spa}

% Generated by code/paper_tables.py; do not edit by hand.
\begin{table}[ht]
\centering
\small
\fittable{lrrrrrrrr}{
\toprule
Target, translator & $\Delta$ & $T$ & zero & characters & meme list & list+analysis & meme-search & given trans. \\
\midrule
Galician, GPT-5.2 & 0.07 & 14 & 0.63\,{\scriptsize$\pm$0.05} & 0.57\,{\scriptsize$\pm$0.03} & 0.56\,{\scriptsize$\pm$0.04} & \textbf{0.35}\,{\scriptsize$\pm$0.06} & -- & 0.03\,{\scriptsize$\pm$0.02} \\
Galician, GPT-6 Sol & 0.30 & 3 & 0.73\,{\scriptsize$\pm$0.30} & 0.51\,{\scriptsize$\pm$0.25} & 0.64\,{\scriptsize$\pm$0.04} & \textbf{0.16}\,{\scriptsize$\pm$0.08} & 0.53\,{\scriptsize$\pm$0.21} & 0.05\,{\scriptsize$\pm$0.04} \\
Spanish, GPT-5.2 & 0.09 & 4 & 0.61\,{\scriptsize$\pm$0.04} & 0.59\,{\scriptsize$\pm$0.04} & 0.57\,{\scriptsize$\pm$0.03} & \textbf{0.23}\,{\scriptsize$\pm$0.11} & -- & 0.07\,{\scriptsize$\pm$0.14} \\
\bottomrule
}
\caption{\textit{Ivan Vasilievich} in the other target languages, evaluator Claude Sonnet 5.5 (automatic evaluation only; human evaluation of GPT-5.2 in Galician: Table \ref{tab:gpt5.2-glg-by-human-mqm}). Galician conditions other than meme-search include the Galician language addendum. MQM adapted score (major-error equivalents per meme; lower is better), mean over translations and evaluation runs, with the 95\% method-level confidence interval. $\Delta$: the smallest difference between two conditions with $T$ translations each that is resolved at the 95\% level, with the variances pooled over the film's conditions; $T$: translations of list+analysis. GPT-5.2: the other conditions have 7 to 21 translations in Galician and 4 to 7 in Spanish; GPT-6 Sol: every condition has $T$ translations. --: GPT-5.2 meme-search was run in English only. Bold: the best condition other than the approved references (given trans.), in each row.}
\label{tab:galician}
\end{table}

Galician and Spanish test whether the pattern holds beyond English. Galician is a lower-resource language with strong interference from Spanish, and an active site for MT evaluation grounded in native-speaker judgment \citep{docampo2024nmtnoninferiority}. All translations were evaluated automatically; the GPT-5.2 Galician translations were also evaluated by the second author, a native speaker (Table \ref{tab:gpt5.2-glg-by-human-mqm}). There is no human evaluation of the GPT-6 Sol Galician or the Spanish translations. In Galician each condition was run with a short language-specific addendum (Supplementary Material, Section~A), and Table \ref{tab:galician} reports these conditions.

The two translators behave differently in Galician (Table \ref{tab:galician}). Unaided, the newer model is no better than the older one on the meme passages, and if anything worse. With only three translations per condition, GPT-6 Sol's differences among the conditions without the analysis are within $\Delta$; given the analysis, it does better than with any of them, better than GPT-5.2, and closes most of the gap to the approved references. The approved references bring both close to the ceiling. In Spanish, where only GPT-5.2 was tested, the analysis helps at least as much as in Galician and more than in English, while the bare meme list again helps no more than generic context (Table \ref{tab:galician}). Across the three target languages, then, the ordering of conditions is the same as in English: identification alone does little, the analysis helps by an amount that varies with language and translator, and the references help most.

% Generated by code/paper_tables.py; do not edit by hand.
\begin{table}[ht]
\centering
\small
\fittable{lrrrrrrr}{
\toprule
Prompt & zero & characters & meme list & list+analysis & given trans. & narratives & added noise \\
\midrule
Without addendum & 0.28\,{\scriptsize$\pm$0.13} & 0.21\,{\scriptsize$\pm$0.03} & 0.44\,{\scriptsize$\pm$0.12} & \textbf{0.15}\,{\scriptsize$\pm$0.08} & 0.00\,{\scriptsize$\pm$0.00} & 0.39\,{\scriptsize$\pm$0.19} & 0.31\,{\scriptsize$\pm$0.11} \\
With addendum & 0.24\,{\scriptsize$\pm$0.05} & 0.36\,{\scriptsize$\pm$0.14} & 0.51\,{\scriptsize$\pm$0.12} & \textbf{0.10}\,{\scriptsize$\pm$0.02} & 0.00\,{\scriptsize$\pm$0.00} & -- & -- \\
\bottomrule
}
\caption{Human evaluation of GPT-5.2, \textit{Ivan Vasilievich}, Russian$\to$Galician, by a native speaker, with and without the Galician language addendum in the prompt. MQM adapted score (major-error equivalents per meme; lower is better), mean over translations, with the 95\% method-level confidence interval. --: condition not run with the addendum. Bold: the best condition other than the approved references (given trans.), in each row.}
\label{tab:gpt5.2-glg-by-human-mqm}
\end{table}

Human and automatic evaluation of the GPT-5.2 Galician translations agree on the ordering of conditions (Table \ref{tab:gpt5.2-glg-by-human-mqm}): approved references reach the ceiling, and the analysis is the best of the conditions without references. As in English, neither the neural metrics, on the whole text or the meme passages alone, nor the whole-text judge reproduce this ordering (Table \ref{tab:metrics-human-gpt52}). The native evaluator is highly self-consistent on repeated judgments (Pearson $r=0.990$, Spearman $\rho=0.933$; $n=8$ repeated items). The language addendum has no consistent effect; the one reliable change is that it makes \textit{characters} worse, plausibly because a formal constraint added to a prompt with little content guidance has room to hurt.

The errors that recur across conditions in Galician are the source-comprehension failures already seen in English. Without the analysis, most conditions render the elliptical line (\ref{ex:doubts-b}) as possession, and leave Bunsha's mock-formal self-introduction as tsar (\ref{ex:tsar}) ambiguous about who is addressing whom, in some translations even giving Bunsha, a man, feminine self-reference (\textit{``Encantada, tsar!''}).

\section{Conclusion}
An AI translator facing a culturally loaded passage has to find it, work out why it is difficult, and only then translate it. Finding the passages is not yet solved: off-the-shelf difficulty estimators rank them no better than chance, and current models searching the web find most of them but not all. Analysis is the bottleneck: the models' own analyses miss the point of some of the memes they find, and on most films help translation less than a human-validated one. Translation with a human-validated analysis is where the model generations differ. The older translator gains modestly from it; the current ones gain on every film, and in human evaluation GPT-6 Sol comes close to the approved references on most. The current translators also gain from merely being told which passages are memes, which the older one did not, and the analysis adds most where a passage hides a concrete trap that it names. Unaided translation of the memes, meanwhile, has barely improved. Models have become much better at using an expert's analysis, and they still need to be given one.

The evaluation protocol is the second contribution. Scoring only the audience-selected passages, against a validated analysis and reference, gives an automatic metric whose ranking of conditions agrees with human judgment. The human judged the passages only, and reference-free neural metrics and a whole-text LLM judge do not agree with it. Part of the reason is dilution: in a whole-text score, the few passages on which translations differ are averaged in with thousands on which they do not. But the neural metrics do not agree on the passages alone either, where they prefer renderings close to the source; what the protocol adds is the selection of passages together with the human-validated reference and analysis the evaluator is given. The evaluator model is a parameter of the protocol: evaluators differ in harshness and in what they overlook (a marked register, a rhyme, an allusion, a legitimate cultural substitution), so absolute scores are comparable only under one evaluator, and the choice of evaluator should be checked against a small human sample. The noise that translators and evaluators add can be measured on a small calibration run, and it determines how many translations and evaluation runs a comparison needs: few where the gain from the analysis is large, many where it is small.

The study covers five films and one source language. The human steps and evaluations were carried out by the authors, one evaluator per target language, so the translations were judged by the people who approved the analyses and references. An independent replication with hired annotators would add inter-annotator agreement, and further source and target languages would test how far the separation of detection from analysis travels. The released code, prompts, and data make both possible without repeating the construction step.

\label{sec:discussion}For a professional translator working with a current LLM, these results suggest a division of labor. The model translates routine dialogue fluently and, given a human-validated analysis, often handles the culturally loaded passages well; what it does not do reliably is decide which passages need that treatment and why. The translator's time is best spent there: finding the passages that carry the text's cultural and comic weight and working out, with the model's help, what each one does. The analyses and references in this study were produced this way, in under one person-day of work per film (\S\ref{sec:hil}). The same procedure---select units by reception, build a validated analysis and reference layer for them, and evaluate at the level of those units---applies to any domain where an audience has already marked which passages matter.

\backmatter

\bmhead{Supplementary information}
\label{sec:app:prompts}
The Supplementary Material gives the full text of every prompt used in this study (Section~A): the shared translation header, the method-specific additions, the Galician language addendum, and both evaluation prompts. Section~B analyzes the complaints of the LLM whole-text judge on the culturally loaded passages.

\bmhead{Acknowledgements}

\section*{Statements and Declarations}

\begin{itemize}
\item \textbf{Funding:} [Funding sources to be confirmed by all authors.]
\item \textbf{Conflict of interest:} The authors declare no competing interests.
\item \textbf{Ethics approval and consent to participate:} This study did not involve external human subjects. All human-in-the-loop analysis, reference-translation, and adequacy-judgment steps were performed by the authors themselves as domain experts (\S\ref{sec:hil}), not by external participants or hired annotators. Ethical approval was not required.
\item \textbf{Consent for publication:} Not applicable.
\item \textbf{Data availability:} The meme dataset, the human-in-the-loop analyses and reference translations, the source subtitles and every full-film translation, all automatic and human evaluations, and all code and prompts are available at \url{https://github.com/olzama/hard-to-find} and archived on Zenodo (\url{https://doi.org/10.5281/zenodo.23504282}). Our annotations, analyses, references, and evaluations are released under CC BY 4.0 and the code under the MIT license; the source subtitles, obtained from OpenSubtitles.org, and their translations are redistributed under the OpenSubtitles.org terms of use.
\item \textbf{Materials availability:} Not applicable.
\item \textbf{Code availability:} See Data availability above.
\item \textbf{Author contribution:} Conceptualization, methodology, software, investigation, formal analysis, data curation, and writing (original draft): Olga Zamaraeva. Investigation (Galician annotation and human evaluation): Carla Martínez. Funding acquisition and writing (review and editing): Carlos Gómez-Rodríguez. All authors read and approved the final manuscript.
\end{itemize}

\bibliography{bilbiography}

\clearpage
\appendix
\setmonofont{FreeMono.otf}
\section*{Supplementary Material}

Section~A gives the full text of every prompt component used in the main paper. Section~B analyzes the complaints of the LLM whole-text judge on the culturally loaded passages.

\section{Prompts}

This appendix gives the full text of every prompt component described in the main paper's Translation Generation and Evaluation sections. All translation prompts share the header below; method-specific material (\S\ref{sec:app:method-prompts}) is appended to it, followed by the language addendum (\S\ref{sec:app:lang-prompt}) where applicable.

\subsubsection{Shared translation header}
\begin{verbatim}
You will perform a film subtitles translation task from {source language}
into {target language}.
{method-specific addition, if any}

Translate the following subtitles in full, even if it takes time, without
stopping midway or asking for clarifications. Preserve time codes EXACTLY.
Return the translation ONLY. Text to translate in full:

{subtitle chunk}
\end{verbatim}

\subsubsection{Method-specific additions}
\label{sec:app:method-prompts}

\paragraph{zero.} Nothing is appended; the shared header above is used as-is.

\paragraph{characters.}
\begin{verbatim}
You will be supplied a list of the main characters of the film, along with
a brief analysis. You should use this additional information to increase
the coherence of the creative structure of the whole translation and to
better identify challenging segments which require additional reasoning
to be translated well.
\end{verbatim}

\paragraph{summary.}
\begin{verbatim}
You will be supplied a summary of the film. You should use it to increase
the coherence of the creative structure of the whole translation and to
better identify challenging segments which require additional reasoning
to be translated well.
\end{verbatim}

\paragraph{narratives.}
\begin{verbatim}
You will be supplied a list of broad cultural narratives contained in the
original text, along with a brief analysis. You should use this additional
information to increase the coherence of the creative structure of the
whole translation and to better identify challenging segments which
require additional reasoning to be translated well.
\end{verbatim}

\paragraph{intertext.}
\begin{verbatim}
You will be supplied a list of intertextual references contained in the
original text, along with a brief analysis. You should use this additional
information to increase the coherence of the creative structure of the
whole translation and to better identify challenging segments which
require additional reasoning to be translated well.
\end{verbatim}

\paragraph{examples.}
\begin{verbatim}
You will be supplied a list of examples of creative/culturally rich/
challenging segments contained in the original, along with a brief
analysis. You should use this additional information to increase the
coherence of the creative structure of the whole translation and to
better identify challenging segments which require additional reasoning
to be translated well.
\end{verbatim}

\paragraph{meme list.}
\begin{verbatim}
You will be provided a list of passages which require special attention
and analysis. You need to perform the analysis yourself: why is the
passage funny or special in the context of the film? How to translate it
effectively? Note: It is not likely that you will succeed without the
analysis, so, make sure to carefully perform it before translating.
\end{verbatim}

\paragraph{list+analysis.}
\begin{verbatim}
You will be provided a list of passages which require special attention
and analysis. You are provided with the approved analysis of why the
passage is funny or special. Use the analysis to translate these specific
passages effectively. Note: It is not likely that you will succeed
without carefully considering the analysis for each item. The analysis
is located in the "analysis" field of the supplied JSON.
\end{verbatim}

\paragraph{given trans.}
\begin{verbatim}
Use the supplied reference JSON file to find reference translations to
use directly for the specified challenging translation items. The JSON
contains the original, the corresponding segment number and the required
translation. Use the reference verbatim and do not add any other
material to the corresponding portion of the translation.
\end{verbatim}

\paragraph{meme-search.}
\begin{verbatim}
1. In the film subtitles you are given, identify passages that have
become "memes" (or similar).
2. Analyze each such meme passage: why it could have become a meme, what
is culturally or pragmatically important/funny about it.
3. When translating that passage, consult the analysis you did above to
achieve a translation that succeeds in creating a similar effect,
whether it is a particular emotion or reference, or something else.
\end{verbatim}

\paragraph{added noise.} Uses the \textit{characters} prompt above, but with the supplied character map swapped for that of a different film and presented as correct (main paper, Translation conditions).

\subsubsection{Language-specific addendum}
\label{sec:app:lang-prompt}
For the ``+lang'' conditions in the extended Galician pilot (main paper), the following is appended after the method-specific material. Galician:
\begin{verbatim}
Language quality requirements for Galician:
- Write native-level Galician as a fluent L1 speaker would.
- Use clitic pronouns correctly according to native Galician usage
  (placement, form, and combinations).
- Avoid castellanismos: do not import Spanish vocabulary, morphology, or
  constructions when a genuine Galician equivalent exists.
- Do not invent words that merely look Galician but that are not
  actually used in the language.
\end{verbatim}

\subsubsection{Evaluation prompts}

\paragraph{Meme-MQM evaluation prompt (main paper, \S~Meme-MQM).}
\begin{verbatim}
You are evaluating film subtitle translations, specifically the segments
which have become memes.

You will receive a JSON object with an array of items. Each item
contains:
- "id": item identifier
- "source": the original text
- "reference": the approved/reference translation
- "analysis": an explanation of the creative properties that the
approved translation captures and of what matters in this item
- "candidate": the actual translation to be evaluated

Task:
Evaluate the CANDIDATE translation for each item against the REFERENCE,
using the ANALYSIS as guidance.

Important principles:
1. Never replace the candidate with the reference.
2. The candidate does NOT need to match the reference literally.
3. If the candidate uses different wording but preserves the key
meaning, tone, humor, voice, register, and artistic effect described in
the ANALYSIS, do not penalize it.
4. If the candidate is identical to the reference, it must receive no
issues.
5. The analysis is not a demand for literal imitation but rather, it
provides a structure for possible alternative reference translations.
Reward equivalent creative solutions.
6. Judge the essence of the analysis: what makes the item work (its
meaning, joke, effect, and the cultural point it makes), not the
specific wording, rhyme words, or particular names, references, or
objects it mentions. A candidate that replaces a name, reference, or
object with a different one that does the same job for the target
audience follows the analysis; it is not a mistranslation.
7. Repetition of a line is NOT an issue if it is repeated in consecutive
segments. This just means the viewer will have more time to read this
line.

You will use the Multidimensional Quality Metrics principles for
translation evaluation. The top level of evaluation will be SEVERITY of
the error/issue.

There are two levels of Severity:
- "major": a problem that significantly harms meaning, humor, style,
character voice, register, historical/epoch feel, or other aspect of
artistic effect; or contradicts the essence of the analysis (see
principle 6). A translation of a joke that does not cross a cultural
boundary is a MAJOR problem.
- "minor": a problem that makes the translation less natural, less
elegant, less precise, or somewhat weaker stylistically, but does not
seriously damage the overall effect. A translation of a joke that is
funny but a bit awkward, is a MINOR problem.

Allowed subcategories, for information only:
- "accuracy"
- "fluency"
- "style"
- "terminology"
- "other"

All categories apply to creative literary translation: what counts is
whether the item works for the target audience, not word-for-word
correspondence.
Use "accuracy" for jokes that are not funny or do not make sense, and
for meaning that is lost or distorted: addition, mistranslation,
omission, or untranslated material. A substitution that does the same
job (principle 6) is not a mistranslation.
Use "fluency" for too literal jokes that are nonetheless funny;
awkwardness, grammar, or poor phrasing.
Use "style" for register, character voice, epoch, tone, or humor/style
mismatch.
Use "terminology" for contextually inappropriate or inconsistent lexical
choices.
Use "other" only if none of the above fits.

Output rules:
- Return one valid JSON object only.
- Do not include markdown.
- The top-level object must contain exactly one key: "items".
- "items" must contain exactly one object per input item, in the same
order.
- Each output item must contain:
  - "id"
  - "issues"
- Each issue must contain:
  - "severity"
  - "category"
  - "justification"
- "justification" must be specific and non-empty.
- If an item has no problems, output `"issues": []`.

Required output format:

{
  "items": [
    {
      "id": 1,
      "issues": [
        {
          "severity": "major",
          "category": "accuracy",
          "justification": "The translation reverses the meaning: the
          Russian implies the items are missing, not that the people
          possess them."
        },
        {
          "severity": "minor",
          "category": "style",
          "justification": "The wording weakens the mock-elevated
          theatrical tone described in the analysis."
        }
      ]
    },
    {
      "id": 2,
      "issues": []
    }
  ]
}
\end{verbatim}

\paragraph{LLM whole-text MQM evaluation prompt (main paper, \S~LLM whole-text MQM).}
\begin{verbatim}
You are evaluating the quality of a film subtitle translation.

You will receive the full source subtitle text and its translation.
There is no reference translation. Your task is to read both texts and
identify translation quality issues anywhere in the translation.

Use the Multidimensional Quality Metrics (MQM) framework with two
severity levels:

- "major": a problem that significantly harms meaning, humor, style,
character voice, register, historical/epoch feel, or other aspect of
artistic effect. A joke that fails to cross the cultural boundary is a
MAJOR problem.
- "minor": a problem that makes the translation less natural, less
elegant, or somewhat weaker stylistically, but does not seriously
damage the overall effect.

Allowed categories:
- "accuracy": mistranslation, omission, addition, or untranslated
material that distorts meaning
- "fluency": unnatural phrasing, grammar errors, awkwardness, or overly
literal rendering
- "style": wrong register, inconsistent character voice, wrong
era/epoch feel, tone mismatch, or humor/style failure
- "terminology": inconsistent or contextually inappropriate lexical
choices

Guidelines:
- This is literary dialogue. Evaluate whether the translation sounds
like natural spoken language in the target language, not like a
transcription.
- Character voices should be consistent throughout. Switches in
register or vocabulary level are worth noting.
- Humor, wordplay, and culturally specific references that fail to land
in the target language are major issues.
- Do not penalize creative solutions that depart from literal meaning
but preserve the artistic effect.
- Do not flag issues for very short subtitle fragments (single words or
exclamations) unless the problem is clear.
- If the translation is generally good, it is acceptable to return few
or no issues.

For each issue, provide:
- "severity": "major" or "minor"
- "category": one of the four above
- "source_quote": the relevant excerpt from the source text (a few
words to a sentence)
- "translation_quote": the corresponding excerpt from the translation
- "justification": a specific explanation of why this is a problem

Output rules:
- Return one valid JSON object only.
- Do not include markdown.
- The top-level object must contain exactly one key: "issues".
- If no issues are found, return {"issues": []}.

Required output format:

{
  "issues": [
    {
      "severity": "major",
      "category": "style",
      "source_quote": "Тебя терзают смутные сомнения",
      "translation_quote": "You are tormented by vague doubts",
      "justification": "The mock-heroic theatrical register of the
Russian is lost; the translation reads as neutral prose rather than
parodic high register."
    },
    {
      "severity": "minor",
      "category": "fluency",
      "source_quote": "Ну что ж",
      "translation_quote": "Well what then",
      "justification": "Awkward phrasing; a natural English equivalent
would be 'Well, so be it' or 'Fair enough'."
    }
  ]
}
\end{verbatim}

\section{The LLM whole-text judge on culturally loaded passages}
\label{sec:app:wf}

To understand the errors of the LLM whole-text judge (GPT-5.4, which judged the GPT-5.2 translations), we pass each flagged item-method pair
to GPT-5.4-mini (temperature 0) for meta-evaluation. For each case where the LLM whole-text judge found
an issue that was matched (via source-quote overlap) to one of the curated reference items (the audience-selected memes of the main paper),
the meta-evaluator receives the Russian source text, the approved reference translation,
the expert analysis, the candidate translation, and the LLM judge's complaint, and is
asked to assess: (1) whether the complaint aligns with what the expert analysis identifies
as the real translation challenge; and (2) whether the complaint is unnecessarily harsh
given the quality of the translation relative to the reference.

\begin{table}[h!]
\centering
\small
\fittable{lrrrrrr}{
\toprule
Film & Items & Issues & Match & Recall & Harsh & Aligned \\
\midrule
\textit{Ivan Vasilievich} & 11 & 573 & 7\% & 73\% & 82\% & 16\% \\
\textit{Diamond Arm} & 23 & 234 & 33\% & 91\% & 80\% & 11\% \\
\textit{Gentlemen of Fortune} & 18 & 262 & 24\% & 78\% & 85\% & 9\% \\
\textit{Carnival Night} & 8 & 263 & 11\% & 88\% & 80\% & 10\% \\
\textit{Pokrov Gates} & 11 & 596 & 9\% & 91\% & 76\% & 9\% \\
\midrule
Total & 71 & 1928 & 14\% & -- & 81\% & 11\% \\
\bottomrule
}
\caption{LLM whole-text judge (GPT-5.4) on the GPT-5.2 translations: behavior on culturally significant passages. Issues: total MQM issues raised across all methods and runs. Match: fraction of issues whose source quote overlapped with a cultural item. Recall: fraction of cultural items flagged in at least one method. Harsh: fraction of flagged item$\times$method pairs rated unnecessarily harsh by meta-evaluator. Aligned: fraction rated as well-aligned with the reference analysis.}
\label{tab:llm-wf-coverage}
\end{table}

The LLM whole-text judge does notice culturally loaded passages, but it spends most of its attention elsewhere: only a small share of its complaints concern a meme passage at all, and an LLM meta-evaluation of those complaints against the approved analyses finds most of them unnecessarily harsh and few aligned with the actual difficulty (Table \ref{tab:llm-wf-coverage}). Three failure modes account for most of the misaligned complaints: the judge penalizes translations that are identical or close to the approved reference; it targets a different aspect of the passage than the one that makes it difficult; and some complaints concern other lines whose source text happens to overlap with the meme.

\end{document}